\documentclass[journal]{IEEEtran}
\usepackage{graphicx}
\usepackage{graphics}
\usepackage{epsfig}
\usepackage{epstopdf}
\usepackage{stfloats}

\usepackage{xcolor,soul,framed} %,caption

\colorlet{shadecolor}{yellow}
\usepackage[cmex10]{amsmath}
\usepackage{array}
\usepackage{mdwmath}
\usepackage{mdwtab}
\usepackage{eqparbox}
\usepackage{url}
\usepackage[colorlinks,
            linkcolor=red, 
             anchorcolor=blue, 
            citecolor=green, 
            ]{hyperref}
\usepackage{float}
\usepackage[ruled,linesnumbered,vlined]{algorithm2e}
\usepackage{algorithmic}
\usepackage{mathrsfs}
\usepackage{amssymb}
\usepackage{amsthm}
\usepackage{cite}
\usepackage{url}
\usepackage{xcolor}
\usepackage{subfig}
\usepackage{fancyhdr}
\usepackage{mdwtab}
\usepackage{caption}
\usepackage{stfloats}
\usepackage{array}
\usepackage{bbding}
\usepackage{subfloat}
\usepackage{verbatim}

\begin{document}
\bstctlcite{IEEEexample:BSTcontrol}
    \title{Flying Bird Object Detection Algorithm in Surveillance Video Based on Motion Information}
  \author{Zi-Wei Sun,
          Ze-Xi Hua,
      Heng-Chao~Li,~\IEEEmembership{Senior Member,~IEEE}, and Hai-Yan Zhong% <-this % stops a space

  \thanks{Finished on June 19, 2023.}
  \thanks{Zi-Wei Sun, Ze-Xi Hua, and Heng-Chao Li are with the School of Information Science and Technology, Southwest JiaoTong University, chengdu 611756, China.}% <-this % stops a space
  \thanks{Hai-Yan Zhong is with the School of Information Science and Technology, Southwest JiaoTong University, chengdu 611756, China, and with Qianghua Times (Chengdu) Technology Co., Ltd., Chengdu 610095, China.}% <-this % stops a space
  }

\markboth{arXiv
}{Ziwei \MakeLowercase{\textit{et al.}}: Flying Bird Object Detection Algorithm in Surveillance Video Based on Motion Information}

% ====================================================================
\maketitle

% === ABSTRACT ====================================================================
% =================================================================================
\begin{abstract}
%\boldmath
A Flying Bird Object Detection algorithm Based on Motion Information (FBOD-BMI) is proposed to solve the problem that the features of the object are not obvious in a single frame, and the size of the object is small (low Signal-to-Noise Ratio (SNR)) in surveillance video. Firstly, a ConvLSTM-PAN model structure is designed to capture suspicious flying bird objects, in which the Convolutional Long and Short Time Memory (ConvLSTM) network aggregated the Spatio-temporal features of the flying bird object on adjacent multi-frame before the input of the model and the Path Aggregation Network (PAN) located the suspicious flying bird objects. Then, an object tracking algorithm is used to track suspicious flying bird objects and calculate their Motion Range (MR). At the same time, the size of the MR of the suspicious flying bird object is adjusted adaptively according to its speed of movement (specifically, if the bird moves slowly, its MR will be expanded according to the speed of the bird to ensure the environmental information needed to detect the flying bird object). Adaptive Spatio-temporal Cubes (ASt-Cubes) of the flying bird objects are generated to ensure that the SNR of the flying bird objects is improved, and the necessary environmental information is retained adaptively. Finally, a LightWeight U-Shape Net (LW-USN) based on ASt-Cubes is designed to detect flying bird objects, which rejects the false detections of the suspicious flying bird objects and returns the position of the real flying bird objects.  The monitoring video including the flying birds is collected in the unattended traction substation as the experimental dataset to verify the performance of the algorithm. The experimental results show that the flying bird object detection method based on motion information proposed in this paper can effectively detect the flying bird object in surveillance video.
\end{abstract}

% === KEYWORDS ====================================================================
% =================================================================================

% \fi
\IEEEpeerreviewmaketitle

% ====================================================================
% ====================================================================
% ====================================================================

% === I. INTRODUCTION =============================================================
% =================================================================================
\section{Introduction}

\begin{figure*}[!ht]
\centering
\includegraphics[width=6.5in]{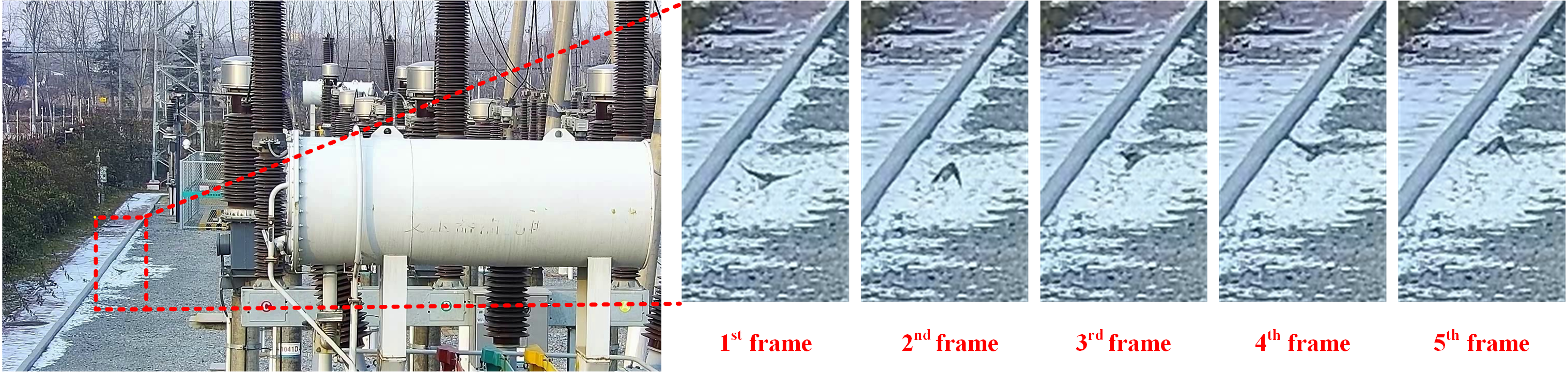}
\caption{On the right, there is a small bird with weak features. Left is a screenshot of the bird on 5 consecutive frames.}
\label{small_and_blurred_bird_fig}
\end{figure*}

\IEEEPARstart{W}{E} are working on the real-time detection of flying bird objects in surveillance videos. For this detection task, there are two main challenges(as shown in Fig. \ref{small_and_blurred_bird_fig} left). 
\begin{itemize}
\item{The features of the object in a single frame are not obvious. Flying birds generally have certain camouflage characteristics similar to the environment, and it is difficult to identify through a single frame.}
\item{The flying bird object is usually small. The monitoring area is generally a room or an outdoor area for general monitoring scenarios. When a bird intrudes into the monitoring area, the proportion of the number of pixels is usually small.}
\end{itemize}

Although the features of the flying bird object in the single frame are not obvious, the flying bird object can still be found by observing the continuous multiple frames of images and using the motion information of the flying bird object (as shown in Fig. \ref{small_and_blurred_bird_fig} on the right). This paper will explore how to use motion information to detect flying bird objects in surveillance videos.

Thanks to the development of deep learning, object detection performance 
  \cite{2014_Girshick_RCNN, 2015_Girshick_Fast_RCNN, 2017_Ren_Faster_RCNN, 2016_Redmon_YOLO, 2016_Liu_SSD, 2017_Redmon_YOLOV2, 2018_Redmon_YOLOv3, 2020_Bochkovskiy_YOLOv4, yolov5_2021, 2021_Zheng_YOLOX} have been significantly improved.  These methods mainly detect objects in still images and rely on the appearance features of the objects. However, the appearance characteristics of flying bird objects are not particularly obvious in surveillance videos, so the object detection method based on still images is ineffective when applied to our task.

In recent years, many methods for video object detection have been proposed. Video has rich temporal information, which we can exploit to enhance the performance of video object detection. It can overcome the problem of the degradation of the detection effect when the video object appears to motion blur, occlusion, and appearance change.

Some video object detection methods take advantage of the fact that the same object may appear in multiple frames within a certain period to improve the robustness of the whole detection. These methods first detect video frames as independent images and then use post-processing methods (such as frame sequence NMS \cite{2016_Han_Seq_NMS}, tracking \cite{2016_Kang_Tubelets_with_CNN}, or other temporal consistency methods \cite{2018_Kang_T-CNN, 2019_Belhassen_Seq-Bbox_Matching}) to enhance the confidence of weak detections in the video. This method requires the algorithm to produce some strong predictions; however, in our task, some flying bird objects are difficult to identify from appearance to disappearance, so we cannot utilize this kind of video object detection method to solve our problem.

Another type of video object detection method uses the idea of feature aggregation \cite{2017_Zhu_DFF, 2017_Zhu_FGFA, 2017_Hetang_Impression_Network, 2018_Zhu_Towards_High_Performance, 2017_Lu_Association_LSTM, 2018_Zhu_Temporally-Aware_Feature_Maps, 2019_Luo_Object, 2019_Wu_SELSA, 2021_Tao_Temporal_RoI_Align, 2022_Han_Class-Aware_Feature_Aggregation, 2022_Xu_Multilevel_Spatial-Temporal_Feature_Aggregation, 2019_Deng_Object_Guided_External_Memory_Network, 2022_Masato_Temporal_feature_enhancement_network_with_external_memory} to improve the overall detection effect. These methods enhance the feature information of the current frame by aggregating the Spatio-temporal information of different frames to improve the detection performance of degraded frames (when the video object has motion blur, occlusion, and appearance change, the detection effect is degraded). However, their main purpose is still to solve the effect degradation problem in video object detection. When applying their idea to our task, it will have some effect (we also need to aggregate the Spatio-temporal information of adjacent frames to enhance the features of the flying bird object in the current frame, thus overcoming the problem of missing object features caused by small or blurred objects in a single frame), but there will still be a lot of false detections and missed detections. We further analyze that their method first extracts the intermediate features of a single frame image and then aggregates the extracted intermediate features to aggregate spatio-temporal information. Since the objects in their experimental data are clear and have large sizes on most images, problems such as short blur, occlusion, and appearance change can be solved after aggregation. However, in our task, the features of most flying bird objects are not obvious on any image, and the size is small. In the feature extraction process of single frame images, the features of the flying bird objects are easily lost, which leads to the failure of spatio-temporal information aggregation. Secondly, due to the object's small size, the problem of low signal-to-noise ratio (unbalanced positive and negative samples) is not easy to eliminate in the training process.

We found that some works detect video objects by generating Spatio-temporal Cubes (St-Cubes) \cite{2017_Rozantsev_Detecting_Flying_Objects} or tubelets \cite{2020_Tang_High_Quality_Object_Linking}. These methods can improve the SNR of video objects, especially small objects; however, since they do not consider that when the object moves too slowly, the object in the pipeline will lack the surrounding environmental information, which is also crucial for our task.

Based on the above analysis, we propose a {\bf{F}}lying {\bf{B}}ird {\bf{O}}bject {\bf{D}}etection algorithm {\bf{B}}ased on {\bf{M}}otion {\bf{I}}nformation (FBOD-BMI) to solve the above problems. Firstly, a ConvLSTM-PAN model structure is designed to capture suspicious flying bird objects, in which the Convolutional Long and Short Time Memory (ConvLSTM) network aggregated the Spatio-temporal features of the flying bird object on adjacent multi-frame before the input of the model and the Path Aggregation Network (PAN) located the suspicious flying bird objects. Then, an object tracking algorithm is used to track suspicious flying bird objects and calculate their Motion Range (MR). At the same time, the size of the MR of the suspicious flying bird object is adjusted adaptively according to its speed of movement (specifically, if the bird moves slowly, its MR wil be expanded according to the speed of the bird to ensure the environmental information needed to detect the flying bird object). Adaptive Spatio-temporal Cubes (ASt-Cubes) of the flying bird objects are generated to ensure that the SNR of the flying bird objects is improved, and the necessary environmental information is retained adaptively. Finally, a LightWeight U-Shape Net (LW-USN) based on ASt-Cubes is designed to detect flying bird objects, which rejects the false detections of the  suspicious flying bird objects and returns the position of the real flying bird objects.

There is an interesting work to detect flying birds, Glances and Stare Detection(GSD)\cite{2019_Tian_GSD}, which simulates the characteristics of human observation of moving small objects by first taking a Glance at the whole image and then taking a closer look at the places that may contain the object. Formally, the region proposal sequence obtained by localization is first used to construct the candidate region sequence, and then the classification method is used to classify the candidate region sequence, and the classification results are fed back to localization to guide the generation of new region proposals. The region proposal is continuously narrowed by alternating localization and classification until the object is locked. The essence of GSD can be seen as solving the problem of low SNR of flying birds. Since the characteristics of the birds in the experimental data set are obvious on a single frame image, this work does not consider the problem that the characteristics of the object are not obvious on a single frame image (the input of the localization stage is the current frame). In this paper, the problem of missing features of the flying bird object on a single frame image and low SNR of the flying bird object is solved simply.

The main contributions of this paper are as follows.

\begin{itemize}
\item{This paper provides an idea to solve the problem that the features of flying bird objects are not obvious in a single frame image, which is to use ConvLSTM to aggregate the spatio-temporal features of flying bird objects on consecutive frames of images before the input of the model, rather than at the intermediate feature layer.}
\item{The ConvLSTM-PAN model structure is designed to capture the suspected flying bird objects. ConvLSTM aggregates the Spatio-temporal features of flying bird objects on consecutive frames, and PAN locates suspicious flying bird objects.}
\item{This paper provides a method to improve the SNR while retaining enough context information for the flying bird objects. Specifically, an adaptive St-Cube extraction method based on the motion information of flying bird objects is proposed in this paper. By using object tracking technology and the motion information of flying bird objects, the St-Cube of the flying bird object in consecutive frames is adaptively adjusted and extracted, so the signal-to-noise ratio of flying bird objects improved, and the necessary environmental information remained.}
\item{A lightweight U-shaped network (LW-USN) model structure was designed to realize the accurate classification and localization of flying bird objects using the St-Cube of suspected flying bird objects.}
\end{itemize}

The remainder of this paper is structured as follows: Section \ref{Related Work} presents work related to this paper. Section \ref{The Proposed Methord} describes the flying bird object detection algorithm based on motion information in detail. In Section \ref{Experiment}, the comparison and ablation experiments of the proposed algorithm are carried out. Section \ref{Conclusion} concludes our work.

% ============================= II. Related Work ==================================
% =================================================================================
\section{Related Work}\label{Related Work}

\subsection{Object Detection in Still Images}\label{MOD_Still_Images}

Traditional object detection methods detect objects by manually designed features related to the object to be detected \cite{2001_Viola_Rapid_object_dection, 2005_Dalal_Histograms, 2010_Felzenszwalb_Object_Detection}. For example, \cite{2005_Dalal_Histograms} proposed the Oriented Gradients (HOG) feature descriptor for feature extraction of object detection. 

Based on the deep learning method, the CNN features related to the object to be detected are automatically extracted by learning to realize the location and classification of the object. For example, the two-stage object detector \cite{2014_Girshick_RCNN, 2015_Girshick_Fast_RCNN, 2017_Ren_Faster_RCNN} first generates region proposals and then uses the CNN features of the region proposals to realize object classification and localization. One-stage object detector \cite{2016_Redmon_YOLO, 2016_Liu_SSD, 2017_Redmon_YOLOV2, 2018_Redmon_YOLOv3, 2020_Bochkovskiy_YOLOv4, 2021_Zheng_YOLOX}, which directly uses an advanced feature extraction network to extract the CNN features of the object for classification and localization. However, they all rely on the appearance feature of the object in the still image, and their detection performance drops sharply when the appearance feature is not obvious.

\subsection{Video Object Detection}\label{MOD_Video}

Compared to still images, video has an additional temporal dimension, which makes it possible for the same object to appear in consecutive video frames. Therefore, the video object possesses redundant and complementary information. Using this redundant and complementary information can improve the performance of video object detection. According to how to use the redundant and complementary information of video objects, video object detection methods can be roughly divided into video object detection methods based on post-processing and video object detection methods based on feature aggregation.

\subsubsection{Video Object Detection with Post-processing}\label{MOD_Video_PP}

These methods first use an image object detector to detect video frames as independent images. Then, the unique Spatio-temporal information of video data is used to improve the accuracy of detection results \cite{2016_Han_Seq_NMS, 2016_Kang_Tubelets_with_CNN, 2018_Kang_T-CNN, 2019_Belhassen_Seq-Bbox_Matching}. For example, Seq-NMS \cite{2016_Han_Seq_NMS} uses Intersection over Union (IOU) to find the same object in adjacent frames and then re-scores the object with low confidence to improve the detection score of the object in the video. Literature \cite{2016_Kang_Tubelets_with_CNN} proposed a complete framework for video object detection based on static image object detection and general object tracking. Firstly, object detection and object tracking techniques are used to generate tubelets object proposal frames. Then, the strong detection boxes in the tubelets are used to enhance the weak detection boxes. Finally, a temporal convolutional network is used to re-score the tubelets. \cite{2018_Kang_T-CNN} extends the popular still image detection framework to solve the generic object detection problem in videos by fusing temporal and contextual information from tubelets. this method effectively incorporates temporal information into the proposed detection framework by locally propagating detection results between adjacent frames and globally modifying detection confidence along the tubelets generated by the tracking algorithm. Literature \cite{2019_Belhassen_Seq-Bbox_Matching} generates tubelets by matching Bboxes across time and then takes a similar idea as Literature \cite{2018_Kang_T-CNN} to correct false and missed detections in the tubelets.

These methods generally require that the detector has better detection performance in some frames or that the object to be detected has rich features in some frames. When the features of the object to be detected are not obvious in any frame, the algorithm will not achieve satisfactory results.

\subsubsection{Video Object Detection Based on Feature Aggregation}\label{MOD_Video_FA}

When the features of the object to be detected are not obvious in some frames, in the feature extraction stage, the feature information of other adjacent frames can be aggregated to enhance the features of the object in these frames. Literature \cite{2017_Zhu_DFF} proposed a method of using optical flow propagation features, then used the features for video object detection, which improved the speed of video object detection. Literature \cite{2017_Zhu_FGFA, 2017_Hetang_Impression_Network, 2018_Zhu_Towards_High_Performance} use optical flow to propagate features and then aggregates the features of the current frame and the propagated features to enhance the expression ability of the current frame to improve the detection accuracy of video objects. Literature \cite{2017_Lu_Association_LSTM, 2018_Zhu_Temporally-Aware_Feature_Maps} use the characteristics of LSTM to realize the aggregation or association of feature information of different frames to realize the detection of video objects. Literature \cite{2019_Luo_Object} models the semantic and spatio-temporal relationships between candidates within the same frame and between adjacent frames to aggregate and enhance the features of each candidate box. In Literature \cite{2019_Wu_SELSA}, the Sequence-level Semantic Aggregation (SELSA) method was proposed, which randomly sampled some frames from the whole video and aggregated the object candidates of the current frame with the weight of the distance between the object candidates to enhance the characteristics of the object candidates of the current frame. Literature \cite{2021_Tao_Temporal_RoI_Align} aggregated the ROI features of the current frame and the most similar ROI features of other frames to obtain the Temporal ROI feature so that the Temporal ROI feature contained the temporal domain information of the object to be detected in the whole video. Literature \cite{2022_Han_Class-Aware_Feature_Aggregation} combined class-aware pixel-level feature fusion and class-aware instance-level feature fusion to improve the detection accuracy of video objects. Literature \cite{2022_Xu_Multilevel_Spatial-Temporal_Feature_Aggregation} introduces a multi-level Spatio-temporal feature aggregation framework, which fully uses frame-level, proposition-level, and instance-level information under a unified framework, to improve the accuracy of video object detection. Literature \cite{2019_Deng_Object_Guided_External_Memory_Network, 2022_Masato_Temporal_feature_enhancement_network_with_external_memory} designs a unique external memory to store the features of other frames and then fuses the features of the current frame to complete feature aggregation.

However, the feature aggregation operation of this kind of method is carried out in the intermediate feature layer.  First, the feature extraction of each frame image is performed separately, and then the extracted features are aggregated, which has a good effect on the ImageNet dataset \cite{2015ImageNet}. However, in our scene, most of the flying birds, the appearance features on any single frame are not particularly rich, the bird size is small, and the bird features are easily lost when extracting the features of a certain frame image alone.  At the same time, looking at consecutive frames makes it more obvious where there are birds, which indicates that the flying bird has complementary feature information on multiple consecutive frames.  Therefore, we use ConvLSTM to aggregate the features of the flying bird object on consecutive frames before the input of the model (instead of aggregating at the intermediate feature layer) and use the aggregated features to locate the flying bird object.

Due to the small proportion of pixels and low SNR of flying birds in the monitoring image, there are still many false and missed detections only using the previous detection method. Therefore, based on the above detection, we use the method of object tracking to obtain the MR of the flying bird object and generate the St-Cubes of the flying bird object to improve the SNR. This St-Cube is then used to classify and localize flying bird objects.

After processing the above methods, the detection effect still does not substantially improve. We further analyzed and found that there was a subset of flying birds with slow movements in our data. At this time, if the St-Cube of the flying bird object is still obtained by the above method, the slow moving flying bird object will lack the surrounding environment information, which will lead to the degradation of the flying bird object detection performance in this part. Therefore, the flying bird object's motion information (motion speed) is considered in this paper. When the motion amount of the flying bird object is lower than a certain threshold, its motion range is expanded, and the environmental information of the flying bird object is increased to improve the identification of the flying bird object.

% === III. The Proposed FBOD-BMI =====
% =================================================================================
\section{The Proposed FBOD-BMI}\label{The Proposed Methord}

\begin{figure*}[!ht]
\centering
\subfloat{\includegraphics[width=6.5in]{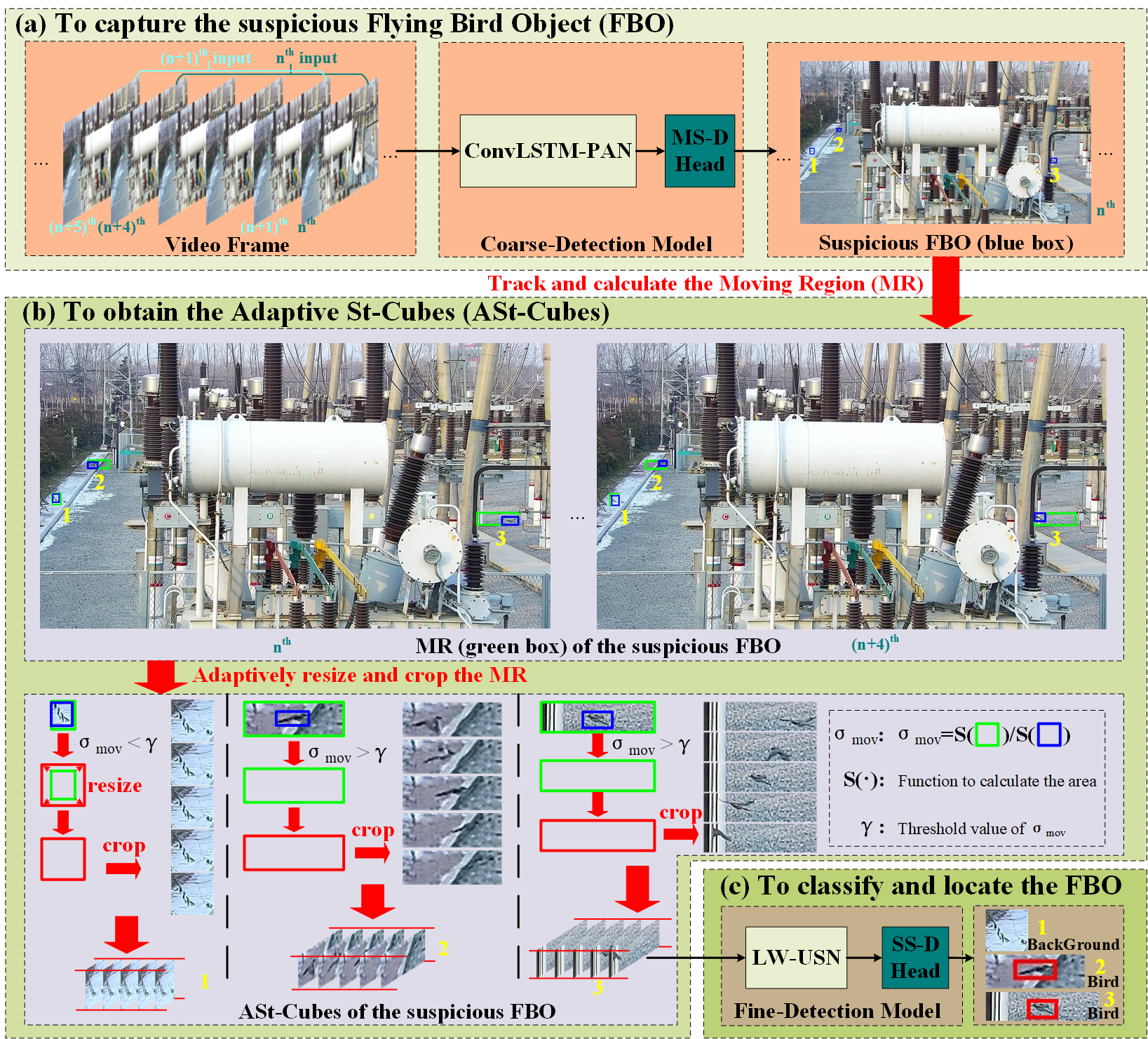}}
\caption{Overview of the proposed FBOD-BMI. (a) Capture the suspicious flying bird object. The blue box in the figure represents the detected suspicious flying bird object. (b) Obtain the ASt-Cubes of the suspicious flying bird object. In the figure, the green box represents the original MR of the flying bird object tracked by the tracking algorithm, and the red box represents the MR adaptively adjusted according to the motion amount of the flying bird object. Since the motion of suspicious flying bird object 1 is less than the set threshold, we expand its MR and increase the environmental information. The amount of motion of suspicious flying bird object 2 and suspicious flying bird object 3 is greater than the set threshold, which indicates that they contain rich environmental information, and we do not change their MR. (c) Classification and localization of flying bird objects.}
\label{framework_fig}
\end{figure*}

Fig. \ref{framework_fig} shows the overview diagram of the proposed FBOD-BMI, which contains three parts. Firstly, ConvLSTM is used to aggregate the features of flying bird objects on consecutive frames before the input of the model, and the aggregated features are used to capture suspicious flying bird objects. Secondly, the adaptive space-time cubes (ASt-Cubes) of suspicious flying bird objects were generated using the method of object tracking and the motion amount of suspicious flying bird objects, which improved the SNR of suspicious flying bird objects while retaining their necessary environmental information. Finally, the ASt-Cubes of the suspicious flying bird object were used to confirm and accurately locate the flying bird object by using a lightweight object detection method. Section \ref{step_a} describes the suspicious flying bird object detection method based on the ConvLSTM feature aggregation technique. Section \ref{step_b} introduces the adaptive Spatio-temporal cube extraction method of suspicious flying bird objects based on object tracking technology and amount of motion. Section \ref{step_c} describes the flying bird object detection method based on ASt-Cubes.

\subsection{To Capture the Suspicious Flying Bird Object}\label{step_a}

In this paper, we perform two steps to capture suspicious flying bird objects (coarse detection of flying bird objects) in consecutive video images. Firstly, the features of flying bird objects on consecutive frames are aggregated. Then, the aggregated features are used to locate the suspicious flying bird object by object detection. This subsection will introduce the acquisition of aggregated features of flying bird objects (Section \ref{step_a_1}) and localization of suspicious flying bird objects (Section \ref{step_a_2}), respectively.

\subsubsection{Acquisition of Aggregated Features for Flying Bird objects}\label{step_a_1}

It has been introduced before that the characteristics of the most flying bird objects are not obvious in a single frame, but they have complementary characteristics in multiple consecutive frames. Therefore, we need to aggregate the features of flying bird objects over multiple consecutive frames.

\begin{figure}[!ht]
\centering
\includegraphics[width=2.6in]{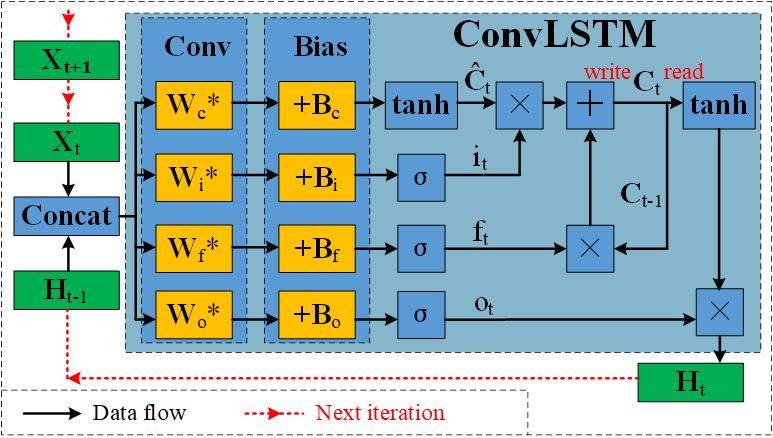}
\caption{Structure diagram of ConvLSTM.}
\label{convlstm_fig}
\end{figure}

The recurrent neural network ConvLSTM(structure shown in Fig. \ref{convlstm_fig}) contains three gates, namely the input gate, output gate, and forget gate, which is used to control the input and output and what information needs to be forgotten and discarded. At the same time, the input gate and output gate can also be understood as controlling the writing and reading of the memory cell. So ConvLSTM can retain useful information and discard redundant or unimportant information.

The coarse-detection phase captures the suspicious flying bird object, and the input is the whole video image, which has the characteristics of many background interference and redundant information (different frames have many identical backgrounds). According to the characteristics of ConvLSTM structure, it can remove redundant or unimportant information while aggregating the features of the flying bird object on consecutive frames. So, in the coarse-detection stage, we use ConvLSTM to extract and aggregate the features of the flying bird object on consecutive frames. Specifically, given the input $n$ consecutive frames of images ${ {{{\bf{X}}_t} \in {{\bf{R}}^{\left( {\text{H} \times \text{W} \times 3} \right)}}|t = \left( {1,2, \cdots ,n} \right)} }$ (Where $\text{H}$ and $\text{W}$ are the height and width of the input image, and $n$ is an odd number), using ConvLSTM network $\mathcal{N}_{\text{ConvLSTM}}$ to extract and aggregate the features of flying bird object on the $n$ consecutive frames, to get spatial-temporal aggregation features ${\bf{H}}_n \in {{\bf{R}}^{\left( {\text{H} \times \text{W} \times \text{C}} \right)}}$ (Where $\text{C}$ is the number of channels) of the flying bird object,
\begin{align}
{\bf{H}}_t = \mathcal{N}_{\text{ConvLSTM}}\left ( \left[ {\bf{X}}_t,{\bf{H}}_{t-1} \right]; {\bf{\Theta}}_{\text{ConvLSTM}} \right ),
\end{align}
where, when $t=1$, ${\bf{H}}_0=\bf{0}$. ${\bf{\Theta}}_{\text{ConvLSTM}}$ is the learnable parameter of the ConvLSTM network. The Spatio-temporal aggregation features ${\bf{H}}_n$ of $n$ consecutive frames of images are input into the subsequent classification and positioning module to determine the category and spatial location information of the suspicious flying bird object.

\subsubsection{Localization of Suspicious flying bird Objects}\label{step_a_2}

\begin{figure}[!ht]
\centering
\includegraphics[width=3.2in]{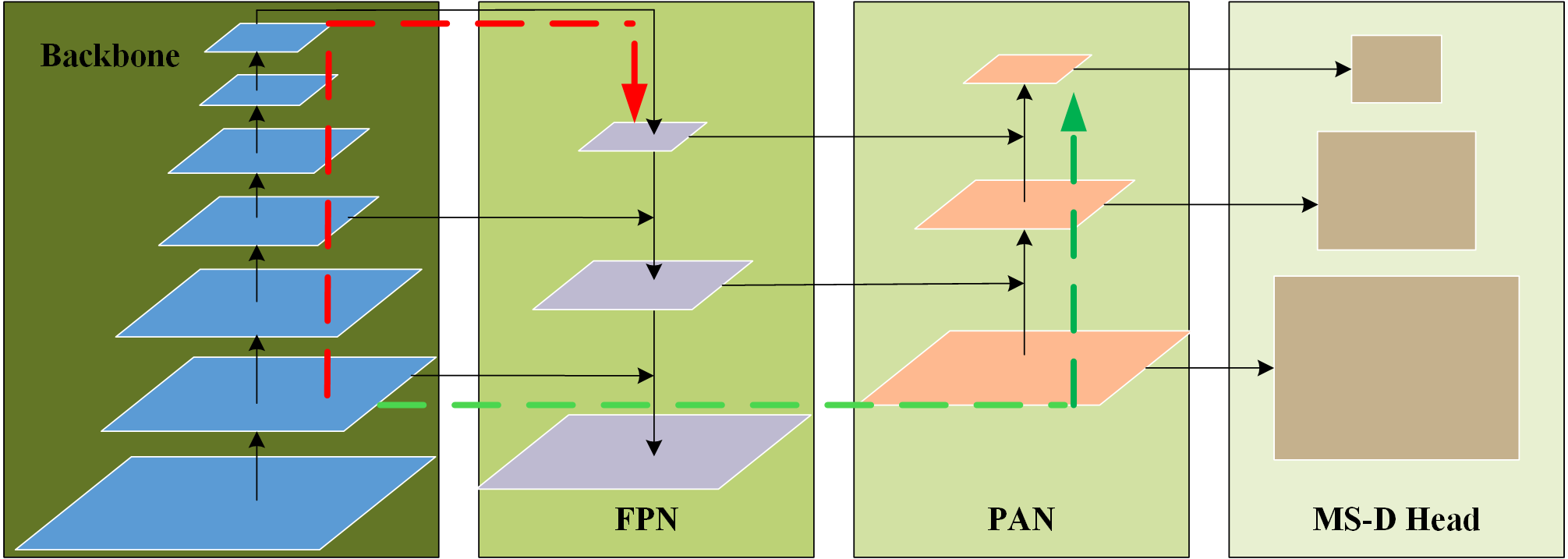}
\caption{Structure diagram of the PANet model}
\label{pan_fig}
\end{figure}

In convolutional neural networks, deeper layers generally have smaller sizes, better global semantic information, and can predict larger objects. The shallower depth layers generally have a larger size, have more delicate spatial information, and can predict smaller objects. However, the large feature layer often does not have a relatively high degree of semantic information, and the small feature layer does not have sufficient spatial positioning information. Therefore, relevant researchers have proposed the structure of FPN \cite{2017_Lin_Feature_Pyramid_Networks} to combine the strong semantic information of the small feature layer and the strong spatial positioning information of the large feature layer. However, the researchers of PANet(Path Aggregation Network) \cite{2018_Liu_Path_Aggregation_Network} found that when FPN transmitted information, there was information loss due to the transfer distance when the information was transmitted to the low-level feature layer. Therefore, path-enhanced FPN, namely PANet structure (see Fig. \ref{pan_fig}), was proposed. The PANet structure opens up a green channel for low-level information transmission and avoids low-level information loss to a certain extent. In this paper, the PANet structure is used to classify and locate flying bird objects.

The Spatio-temporal aggregation features ${\bf{H}}_n$ of $n$ consecutive frames are input into the PANet structure to extract high-level abstract features of flying bird objects ${\bf{F}}_{{\text{FBO}}_n}$,
\begin{align}
{\bf{F}}_{{\text{FBO}}_n} = \mathcal{N}_{\text{PAN}}\left ( {\bf{H}}_n; {\bf{\Theta}}_{\text{PAN}} \right ),
\end{align}
where, ${\bf{\Theta}}_{\text{PAN}}$ is the learnable parameter of the PANet network. The high-level abstract features ${\bf{F}}_{{\text{FBO}}_n}$ of flying bird objects can be directly used for classification and localization of flying bird objects.

When the distance between the flying bird object and the surveillance camera is different, the size of the flying bird object is also different, so the flying bird object to be detected has a multi-scale property. According to the multi-scale property of the flying bird object, this paper uses the MultiScale Detection Head (MS-D Head) to detect the suspicious flying bird object.

The objects in the middle frame of $n$ consecutive frames have symmetric contextual information, which can get more accurate results in prediction. Therefore, this paper predicts the suspicious flying bird object in the middle frame of $n$ consecutive frames as the detection result of the Coarse-detection stage. Specifically, the high-level abstract feature ${\bf{F}}_{{\text{FBO}}_n}$ of the flying bird object is input into the MS-D Head to obtain the output of the model,
\begin{align}
{\bf{O}}_n = \mathcal{N}_{\text{MS-D}}\left ( {\bf{F}}_{{\text{FBO}}_n}; {\bf{\Theta}}_{\text{MS-D}} \right ),
\end{align}
where, ${\bf{\Theta}}_{\text{MS-D}}$ is the learnable parameter of the MS-D Head. Then post-processing operations such as Boxes Decoding and non-maximum suppression were performed on the output of the model to obtain the location of the suspicious flying bird object in the middle frame of $n$ consecutive frames,
\begin{align}
{\left\{ {\text{P}_{\text{ID1}}}, \cdots, {\text{P}_{\text{IDk}}} \right\}}_{\text{frame}^{\left( \frac{n+1}{2}\right)}}= \mathbb{F}_{\text{P}}\left ( {\bf{O}}_n \right ),
\end{align}
where, ${\left\{ \cdot \right\}}_{\text{frame}^{\left( \frac{n+1}{2}\right)}}$ means the locations of the flying bird objects in $\left( \frac{n+1}{2}\right )^{\text{th}}$ frames. ${\text{P}_{\text{IDk}}}$ indicates the predicted position of the object with $\text{IDk}$ (the object with $\text{IDk}$ is taken as an example unless otherwise specified). $\mathbb{F}_{\text{P}}\left ( \cdot \right )$ denotes the post-processing method.

\subsection{To Obtain the ASt-Cubes of flying bird objects}\label{step_b}

In this paper, we extract the St-Cubes of the suspicious flying bird objects on $n$ consecutive frames to improve the SNR of the flying bird objects. At the same time, to ensure the necessary environmental information of the suspicious flying bird objects, the size of the MR is adaptively adjusted according to the motion amount of the suspicious flying bird so that the subsequent detection results are more accurate. Specifically, we will divide it into two steps to obtain the ASt-Cubes of suspicious flying bird objects. Respectively, the original MR of the suspicious flying bird object is extracted using the object tracking technology (Section \ref{step_b_1}), and the MR is adaptively adjusted using the motion amount of the suspicious flying bird object to obtain the ASt-Cubes (Section \ref{step_b_2}).

\subsubsection{Acquisition of the Original MR of the Suspicious flying bird Object}\label{step_b_1}

\begin{figure*}[!ht]
\centering
\includegraphics[width=5in]{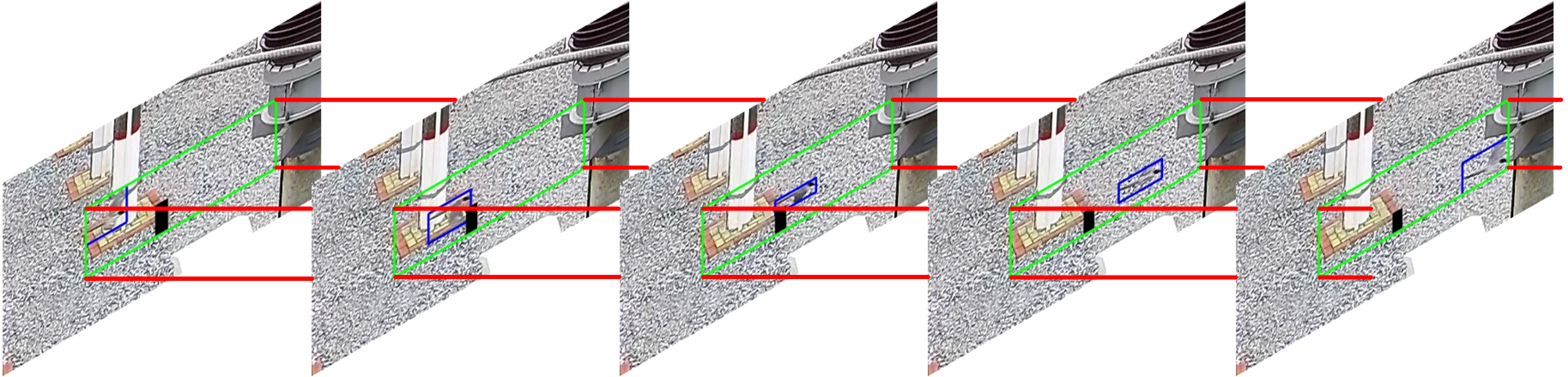}
\caption{MR of the flying bird over five consecutive frames. The blue box shows the bird's position in each frame; The green box represents the minimum bounding rectangle of the five blue boxes, which is the MR of the flying bird over five consecutive frames.}
\label{MR_fig}
\end{figure*}

From the $\left( \frac{n+1}{2}\right )^{\text{th}}$ frame, there are detection results of the suspicious flying bird object, and we start to track the suspicious flying bird object from the $\left( \frac{n+1}{2} + 1\right )^{\text{th}}$ frame. In some cases, the appearance characteristics of flying bird objects are not obvious, so we only use their motion information when tracking them and use a relatively simple SORT \cite{2016_Bewley_SORT} object tracking algorithm to track suspicious flying bird objects,
\begin{align}
\left\{ {\left\{ {\text{P}_{\text{IDk}}}\right\}}_{\text{frame}^{\left( i \right)}}, {\left\{ {\text{P}_{\text{IDk}}}\right\}}_{\text{frame}^{\left( i+1 \right)}}, \cdots\right\}= \mathbb{F}_{\text{track}}\left ( \text{IDk} \right ),
\end{align}
where, $\left\{ {\left\{ {\text{P}_{\text{IDk}}}\right\}}_{\text{frame}^{\left( i \right)}}, {\left\{ {\text{P}_{\text{IDk}}}\right\}}_{\text{frame}^{\left( i+1 \right)}}, \cdots\right\}$ represents the position on consecutive image frames of a suspicious flying bird object with $\text{ID}$ number $\text{k}$, and $\mathbb{F}_{\text{track}}\left ( \cdot \right )$ represents the SORT object tracking method. After obtaining the position of the suspicious flying bird object on consecutive image frames, we can find the Motion Range (MR) of the suspicious flying bird object on $n$ consecutive frames. Specifically, the minimum circumscribed rectangle ${\text{Rect}}_{\text{IDk}}$ at $n$ positions is calculated according to the position of the same object on $n$ consecutive frames of images,
\begin{align}
\begin{split}
&{\text{Rect}}_{\text{IDk}} = \\
&\mathbb{F}_{\text{MinRect}}\left ( \left\{ {\left\{ {\text{P}_{\text{IDk}}}\right\}}_{\text{frame}^{\left( i+1 \right)}}, \cdots, {\left\{ {\text{P}_{\text{IDk}}}\right\}}_{\text{frame}^{\left( i+n \right)}}\right\} \right ),
\end{split}
\end{align}
where, $\mathbb{F}_{\text{MinRect}}\left (\cdot \right )$ denotes the function to find the minimum circumscribed rectangle of $n$ rectangular boxes. For example, to find the minimum circumscribed rectangle  $\left[ \left( x_\text{min}, y_\text{min} \right), \left( x_\text{max}, y_\text{max} \right) \right]$ (Using the horizontal and vertical coordinates of the top left and bottom right vertices of the rectangle) of $\left\{ {box}_1, \cdots, {box}_n \right\}$, the specific calculation method is as follows,
\begin{align}
\begin{split}
x_\text{min}&=\text{min}\left( x_{1_{{box}_1}}, \cdots, x_{1_{{box}_n}}\right),\\
y_\text{min}&=\text{min}\left( y_{1_{{box}_1}}, \cdots, y_{1_{{box}_n}}\right),\\
x_\text{max}&=\text{max}\left( x_{2_{{box}_1}}, \cdots, x_{2_{{box}_n}}\right),\\
y_\text{max}&=\text{max}\left( y_{2_{{box}_1}}, \cdots, y_{2_{{box}_n}}\right),
\end{split}
\end{align}
where, $\left[ \left( x_{1_{{box}_n}}, y_{1_{{box}_n}} \right), \left( x_{2_{{box}_n}}, y_{2_{{box}_n}} \right) \right]$ denotes the horizontal and vertical coordinates of the upper left and lower right vertices of ${box}_n$ in the image. The obtained minimum circumscribed rectangle ${\text{Rect}}_{\text{IDk}}$ is the MR of the flying bird object in $n$ consecutive frames. The diagram of the motion range of the flying bird object on five consecutive frames of images is shown in Fig. \ref{MR_fig}, which, after cropping, can be used as a St-Cubes for flying bird objects.

\subsubsection{Adaptively Adjust the MR to Obtain ASt-Cubes Based on the Amount of Motion}\label{step_b_2}

\begin{figure*}[!ht]
\centering
\includegraphics[width=5in]{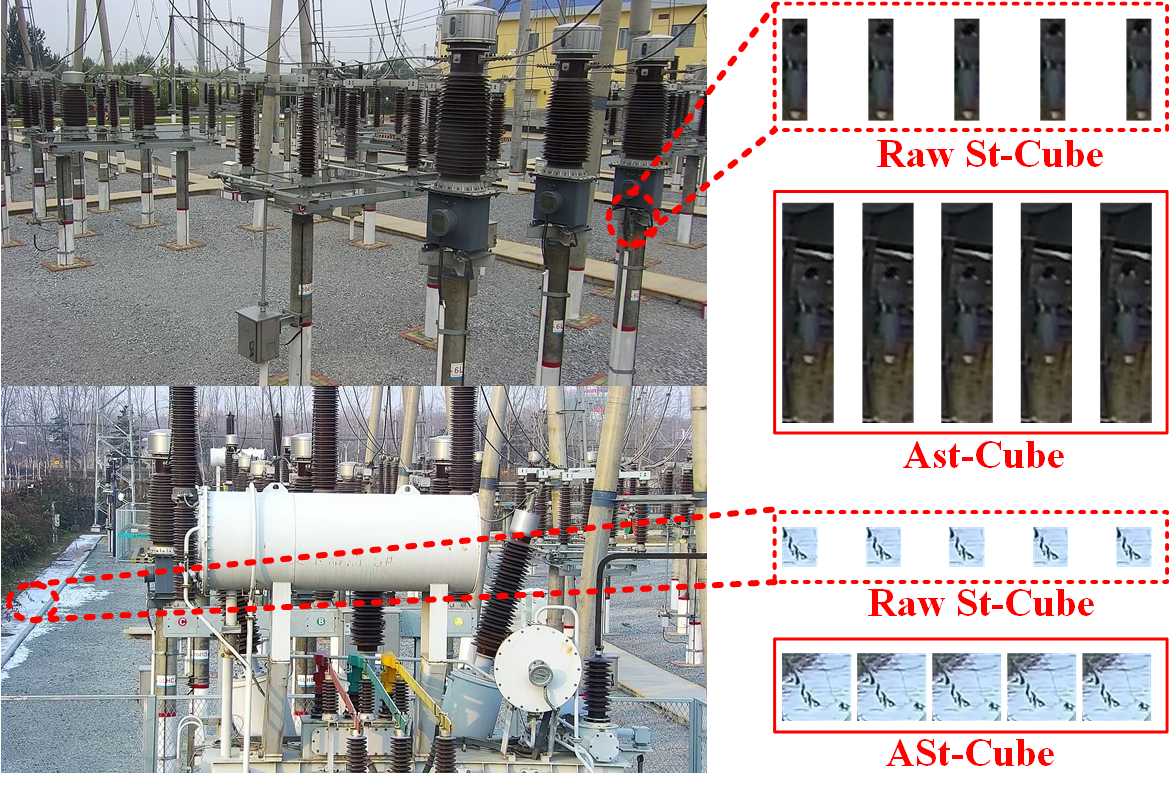}
\caption{The left figure is the original monitoring picture, the right figure is the St-Cubes expansion diagram of the flying bird object in five consecutive frames in the dotted frame, and the solid frame is the St-Cubes adaptively adjusted according to the amount of object motion. Obviously, the object in the original St-Cubes is difficult to recognize correctly, and the object in the ASt-Cubes is easier to recognize.}
\label{Raw_SC_VS_ASC_fig}
\end{figure*}

We crop the MR of the suspicious flying bird object in $n$ consecutive frames to generate the St-Cubes of the flying bird object. The St-Cube eliminates the interference of other background and negative samples, which can improve the SNR of the flying bird object. However, suppose the flying bird object moves too slowly. In that case, the resulting St-Cubes will lack the necessary environmental information (see Raw St-Cube in Fig. \ref{Raw_SC_VS_ASC_fig}), which is not conducive to the detection of flying bird objects. To balance the contradiction between SNR and environmental information, this paper proposes an ASt-Cubes extraction method based on the amount of motion of the flying bird object, which adaptively adjusts the size of the MR of the flying bird object according to the speed of the object's motion. There are two steps.

Firstly, the amount of motion of the flying bird object over $n$ consecutive frames is calculated. If an object of the same size moves fast on $n$ consecutive frames, its MR is large; otherwise, its MR is small. Therefore, we use the ratio of the area of the MR of the flying bird object on $n$ consecutive frames to the area of the single frame image occupied by the flying bird object to define its motion amount on $n$ consecutive frames,
\begin{align}
{\sigma}_{mov}= \frac{S\left( {\text{Rect}}_{\text{IDk}} \right)}{S\left( {\text{Obj}}_{\text{IDk}} \right)},
\end{align}
where, ${\sigma}_{mov}$ is the motion amount, $S\left( \cdot \right)$ represents the function to calculate the area, and ${\text{Obj}}_{\text{IDk}}$ represents the object with $\text{ID}$ number $\text{k}$. The area of MR is then the area of the minimum circumscribed rectangle ${\text{Rect}}_{\text{IDk}}$. Since the area occupied by a flying bird object in a single image frame may vary due to its shape changes, and it is not easy to calculate accurately, we use the rectangular area of its bounding box to approximately represent its area in this paper.

Then, according to the amount of motion of the flying bird object, the MR of the flying bird object is adaptively adjusted as the Adaptive MR ${\text{ARect}}_{\text{IDk}}$ of the flying bird object. Specifically, a motion hyperparameter $\gamma$ is introduced. When the amount of motion of the flying bird object is less than $\gamma$, the MR of the flying bird object is expanded to make the amount of motion of the flying bird object reach $\gamma$. Therefore, the Adaptive MR of the flying bird object can be expressed as follows,
\begin{align}
{\text{ARect}}_{\text{IDk}}=
\begin{cases}
     {\text{ARect}}_{\text{IDk}}, &{\sigma}_{mov} \geq \gamma\\
    \gamma \times {\text{Obj}}_{\text{IDk}}, &{\text {otherwise}}.
\end{cases}
\end{align}
The ${\text{ARect}}_{\text{IDk}}$ is used to crop the corresponding $n$ consecutive frames of video image $\left\{ {\text{frame}^{\left( 1 \right)}}, \cdots,  {\text{frame}^{\left( n \right)}} \right\}$ respectively, and the $n$ frame screenshots obtained are the Adaptive St-Cubes (ASt-Cubes) (${\text{ASC}}_{\text{IDk}}$) of the flying bird object,
\begin{align}
 \text{f}^{\left( i \right)}_{{\text{ARect}}_{\text{IDk}}}&=  \mathbb{F}_{\text{cut}}\left ( {\text{frame}^{\left( i \right)}}, {\text{ARect}}_{\text{IDk}} \right ),\\
 {\text{ASC}}_{\text{IDk}}&= \left\{  \text{f}^{\left( i \right)}_{{\text{ARect}}_{\text{IDk}}} \vert i \in \left( 1, \cdots,  n \right) \right\}.
\end{align}

\subsection{flying bird Object Detection Based on St-Cube}\label{step_c}

After the previous processing, we improve the SNR of the flying bird object, retain its necessary environmental information, and obtain the ASt-Cubes of the flying bird object. In the fine-detection stage, we can use the ASt-Cubes of the flying bird object to classify and locate the flying bird object. Specifically, the fine-detection phase includes feature aggregation of flying bird objects (\ref{step_c_1}) and flying bird objects' classification and localization (\ref{step_c_2}).

\subsubsection{feature aggregation of flying bird objects}\label{step_c_1}

The input of the fine-detection model is the ASt-Cubes of the flying bird object extracted earlier. The coarse-detection model may detect multiple suspicious flying bird objects at one time, so there will be multiple ASt-Cubes, and the fine-detection model will detect each ASt-Cubes separately. So it is possible to run a coarse-detection model once and a fine-detection model many times. Therefore, to balance the accuracy and speed, the way of aggregating the Spatio-temporal features of the flying bird object in the fine-inspection stage is used to merge the input of multiple consecutive frames. At the same time, to reduce data redundancy, except for the middle frame, the rest of the frames are input in the form of a grayscale image single channel. Specifically, firstly, grayscale the screenshots of the ASt-Cubes of the flying bird object except for the middle screenshot,
\begin{align}
 \text{f}^{{\left( i \right)}^{\prime}}_{{\text{ARect}}_{\text{IDk}}}=
 \begin{cases}
     \text{f}^{\left( i \right)}_{{\text{ARect}}_{\text{IDk}}}, &\text{i} = \text{int}\left( \frac{n}{2}\right)\\
     \mathbb{F}_{\text{Gray}}\left( \text{f}^{\left( i \right)}_{{\text{ARect}}_{\text{IDk}}} \right), &{\text {otherwise}}.
\end{cases}
\end{align}
where, $ \mathbb{F}_{\text{Gray}}\left ( \cdot \right )$ is a function that finds the grayscale of a color image. Then, the processed screenshots of the ASt-Cubes are Concatenate in the channel dimension as the input of the fine-detection stage,
\begin{align}
{\bf{X}}_{{\text{S}}_{\text{IDK}}}=  \mathbb{F}_{\text{concat}}\left ( \left( \text{f}^{{\left( 1 \right)}^{\prime}}_{{\text{ARect}}_{\text{IDk}}}, \cdots,  \text{f}^{{\left( n \right)}^{\prime}}_{{\text{ARect}}_{\text{IDk}}} \right), 2 \right ),
\end{align}
where, the second argument of the $\mathbb{F}_{\text{concat}}$ function indicates that the concatenation operation is performed in the third input dimension (height, width, channel). The length and width of ${\bf{X}}_{{\text{S}}_{\text{IDK}}}$ are equal to the length and width of rectangle ${{\text{ARect}}_{\text{IDk}}}$, and the number of channels is $n+2$, which contains the flying bird object's features in the ASt-Cubes. It is input into the fine-detection model to classify and locate the flying bird object accurately.

\subsubsection{Classification and Localization of flying bird objects}\label{step_c_2}
\begin{figure}[!ht]
\centering
\includegraphics[width=2in]{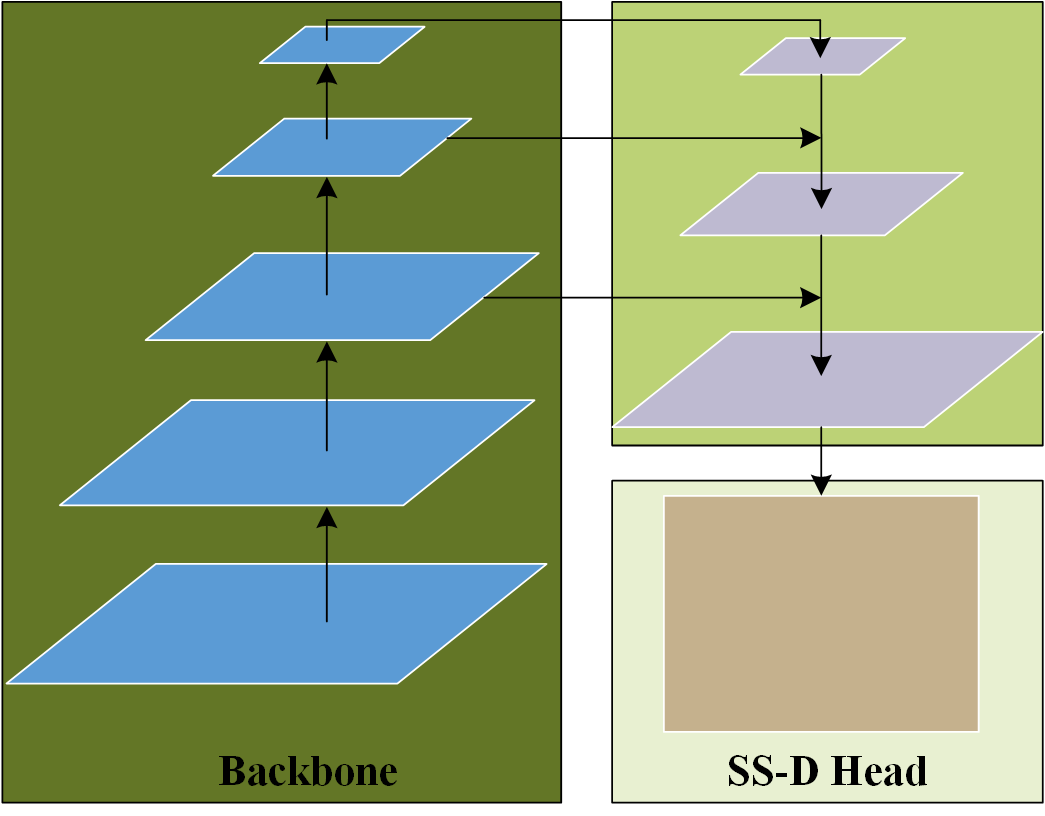}
\caption{Structure diagram of LW-USN}
\label{usnet_fig}
\end{figure}

In order to further improve the speed of the whole flying bird object detection process, this paper uses a lightweight U-Shaped Network (USN) (in the experiment, we use MobilenetV2 
 \cite{2018_Sandler_MobileNetV2}as the backbone network of the USN) as the feature extraction network of the flying bird object in the fine-detection stage, as shown in Fig. \ref{usnet_fig}.

We feed ${\bf{X}}_{{\text{S}}_{\text{IDK}}}$, which aggregates the ASt-Cubes of the flying bird object, into the LW-USN feature extraction network to obtain the aggregated ASt-Cubes feature ${\bf{F}}_{\text{IDK}}$ of the flying bird Object,

\begin{align}
{\bf{F}}_{\text{IDK}} = \mathcal{N}_{\text{LW-USN}}\left ( {\bf{X}}_{{\text{S}}_{\text{IDK}}}; {\bf{\Theta}}_{\text{LW-USN}} \right ),
\end{align}
where, ${\bf{\Theta}}_{\text{LW-USN}}$ is the learnable parameter of the LW-USN.

The ASt-Cube of a flying bird object may contain more than one object. Moreover, due to the interference of background and negative samples, the detection accuracy of the coarse-detection model is not satisfactory; there will be false and missed detection. So, the ASt-Cube may contain no object, one object, or multiple objects. Therefore, the detection model in the fine-detection stage should still have the ability of multi-object detection. However, since the ASt-Cubes of flying bird objects are only a small area (relative to the input image) and cannot contain many flying bird objects, the output of the fine-detection model need not be designed with a complex structure. In summary, the paper uses a relatively simple Single Scale Detection Head (SS-D Head) structure as the output structure of the fine-detection model (see Fig. \ref{usnet_fig}). Specifically, ${\bf{F}}_{\text{IDK}}$ is fed into the SS-D Head to obtain the output of the fine-detection model,
\begin{align}
{\bf{O}}_{\text{IDK}} = \mathcal{N}_{\text{SS-D}}\left ( {\bf{F}}_{\text{IDK}}; {\bf{\Theta}}_{\text{SS-D}} \right ),
\end{align}
where, ${\bf{\Theta}}_{\text{SS-D}}$ is the learnable parameter of the SS-D Head. Then, post-processing operations such as Boxes Decoding and non-maximum suppression are performed on the output to obtain the final detection result of the flying bird object,
\begin{align}
\left\{{Classes}_{\text{IDK}},{Boxes}_{\text{IDK}} \right\}  = \mathbb{F}_{\text{P}}\left ( {\bf{O}}_{\text{IDK}} \right ),
\end{align}
where, ${Classes}_{\text{IDK}}$ represents the category of the object in the ASt-Cube ${\text{ASC}}_{\text{IDk}}$ of the flying bird object, and ${Boxes}_{\text{IDK}}$ (in this paper, the position of the object in the middle frame of $n$ consecutive frames is taken as the detection result) is the bounding box of the corresponding object in this region.

Finally, the bounding box of the flying bird object in the ASt-Cubes is mapped to the original video image. That is, the final detection result of the flying bird object is obtained.

% === IV. The Experiment ========================================
% =================================================================================
\section{Experiment}\label{Experiment}
In this section,  A series of experiments are conducted to quantitatively and qualitatively evaluate the proposed FBOD-BMI. Next, we will introduce datasets (\ref{datasets}), evaluation metrics (\ref{evaluation_metrics}), experimental platforms (\ref{experimental_platforms}), implementation details (\ref{implementation_details}), and parameter analysis experiment (\ref{parameter_analysis}), comparative analysis experiment (\ref{comparative_analysis}).

\subsection{Datasets}\label{datasets}

We collected and annotated 115 videos containing flying birds (the size of video images is 1280 $\times$ 720) in an unattended traction substation. We end up with 28156 images, of which 7634 contain flying birds, with 8589 flying bird objects. From the 115 videos, we randomly selected 18 videos for testing and the remaining 97 videos for training. The training set includes 24771 images, and the test set includes 3385 images.

\begin{figure*}[!ht]
\centering
\includegraphics[width=7in]{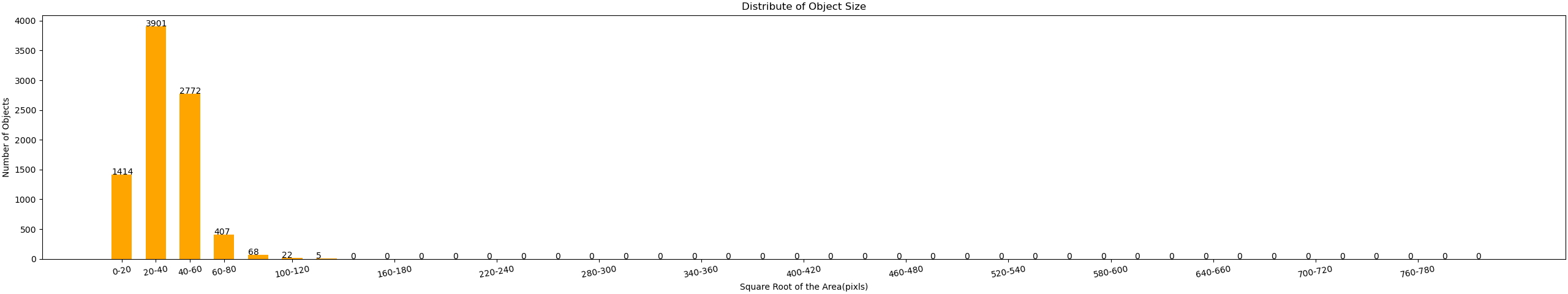}
\caption{Size distribution of the flying birds in the datasets}
\label{size_curve_fig}
\end{figure*}

From Fig. \ref{size_curve_fig}, we can see that the size of flying birds is mainly distributed between 0 $\times$ 0 and 80 $\times$ 80 pixels, and more than 60\% are below 40 $\times$ 40 pixels. 

\subsection{Evaluation Metrics}\label{evaluation_metrics}

In this paper, the widely used measures in object detection, average precision (AP), is adopted to evaluate the proposed FBOD-BMI. More specifically, $\text{AP}_{50}$ (The subscript 50 means that the detection result is regarded as the True Positive when the IOU between the detection result and the ground truth is greater than or equal to 50\%.  That is, the IOU threshold is set 50\% ), $\text{AP}_{75}$ (The subscript 75 has a similar meaning with the subscript 50), and AP (Average Precision averaged over multiple thresholds, IOU threshold is set from 50\% to 95\%, in intervals of 5\%) are adopted.

\subsection{Experimental Platforms}\label{experimental_platforms}

All the experiments are implemented on a desktop computer with an Intel Core i7-9700 CPU, 32 GB of memory, and a single NVIDIA GeForce RTX 3090 with 24 GB GPU memory.

\subsection{Implementation Details}\label{implementation_details}

We implemented the proposed method based on YOLOV4 \cite{2020_Bochkovskiy_YOLOv4} with modifications.

Specifically, for the coarse-detection model, we directly place the ConvLSTM module before the input of the CSPDarkNet53, the backbone network of the YOLOV4 model (i.e., the input of CSPDarkNet53 was originally an image, but now the input is the output of ConvLSTM aggregating the features of $n$ consecutive frames). As shown in Fig. \ref{Aggregation mode 1}. For the input size of the coarse-detection model, we set it to 640 $\times$ 384 to ensure the ratio of effective input pixels as much as possible and, at the same time, ensure the running speed. During training, the input is $n$ consecutive frames of images, the label is the object's position on the intermediate frame, and the loss function of the YOLOV4 algorithm is reused.

For the fine-detection model, the lightweight MobilenetV2 is used as the backbone of the U-shaped network. For the input size of the fine-detection model, we set it to 160 $\times$ 160. For the training data, we used the coarse-detection model and the object tracking SORT algorithm to collect the Motion Region (MR) containing the flying bird object as the positive and negative samples without the object. During training, the input is the screenshot of the MR of $n$ consecutive frames, the label is the object's position on the intermediate screenshot, and the loss function of YOLOV4 is reused.

In this paper, all experiments are implemented under the Pytorch framework. All network models are trained on an NVIDIA GeForce RTX 3090 with 24G video memory. The batch size setting is set to 4 when training the coarse-detection model designed in this paper and other comparable models, and it is set to 8 when training the fine-detection model. All the experimental models were trained from scratch, and no pre-trained models were used. The trainable parameters of the network were randomly initialized using a normal distribution with a mean of 0 and a variance of 0.01. Adam was chosen as the optimizer for the model in this paper. The initial learning rate is set to 0.001. For each iteration, the learning rate is multiplied by 0.95, and the model is trained for 100 iterations. In the training phase, we used simple data augmentation, including random horizontal flipping, random Gaussian noise, etc., to enhance the robustness of the model.

\subsection{Parameter Analysis Experiments}\label{parameter_analysis}

\subsubsection{Effect of feature aggregation at different locations on the accuracy of the algorithm}

\begin{figure}[htb]
\centering
\subfloat[Original backbone of CSPDarkNet53]{\includegraphics[width=2.906in]{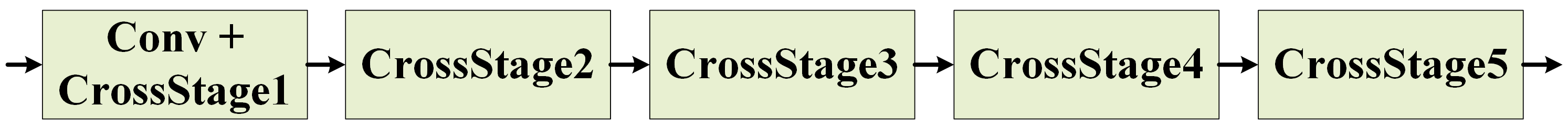}
\label{Original backbone of CSPDarkNet53}}
\\
\subfloat[Aggregation mode I]{\includegraphics[width=3.2in]{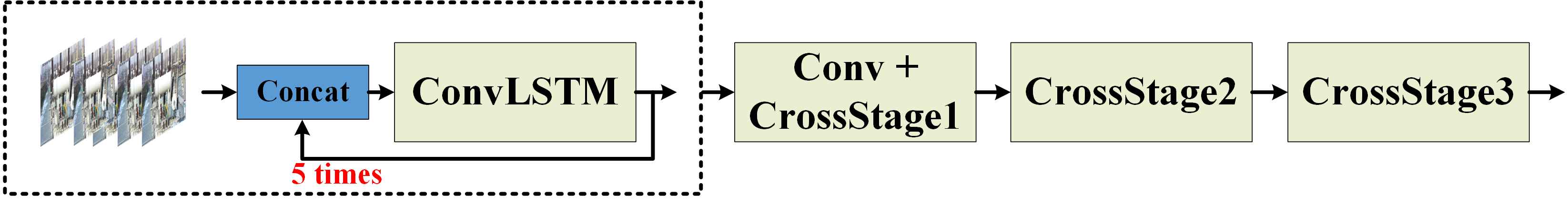}
\label{Aggregation mode 1}}
\\
\subfloat[Aggregation mode II]{\includegraphics[width=3.2in]{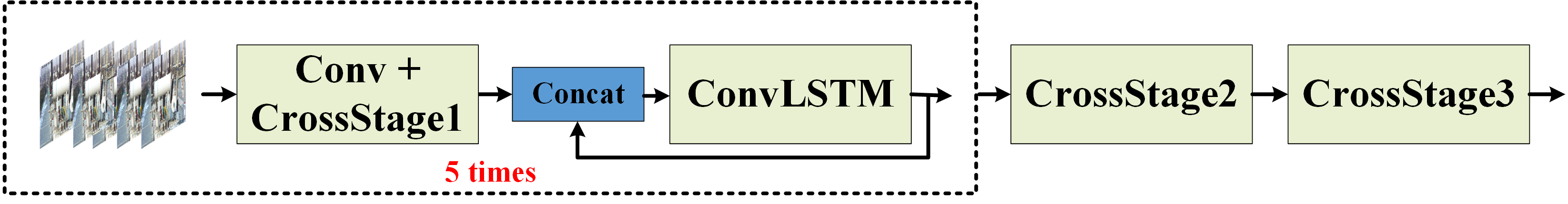}
\label{Aggregation mode 2}}
\\
\subfloat[Aggregation mode III]{\includegraphics[width=3.2in]{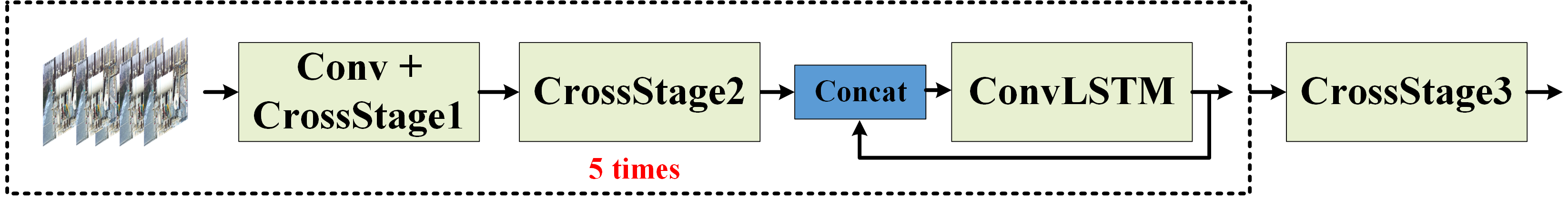}
\label{Aggregation mode 3}}
\caption{Aggregation mode. Fig. \ref{Original backbone of CSPDarkNet53} shows the original backbone network. Fig. \ref{Aggregation mode 1} Aggregation operation is before the input of the backbone network. Fig. \ref{Aggregation mode 2} Aggregation operation is between the first CrossStage module and the second CrossStage module of the backbone network. Fig. \ref{Aggregation mode 3} Aggregation operation is between the second CrossStage module and the third CrossStage module of the backbone network.}
\label{Aggregation mode}
\end{figure}

To prove that, for the characteristics of flying birds in our task (most flying birds do not have obvious features on a single frame image), it is necessary to perform feature aggregation on continuous video images before input to the model, and we set up a set of comparison experiments. Specifically, we set 3 modes for aggregating features of flying birds. In the first (aggregation mode I), we directly place the ConvLSTM module before the input of the CSPDarkNet53, the backbone network of the YOLOV4 model (i.e., the input of CSPDarkNet53 was originally an image, but now the input is the output of ConvLSTM aggregations of the features of five consecutive frames of images. As shown in Fig. \ref{Aggregation mode 1}); In the second (aggregation mode II), the ConvLSTM module is embedded between the first CrossStage module and the second CrossStage module of CSPDarkNet53, the backbone network of YOLOV4 model (that is, for each image frame, Need to run to the first CrossStage module of CSPDarkNet53 to perform feature aggregation, as shown in Fig. \ref{Aggregation mode 2}); In the third (aggregation mode III), the ConvLSTM module is embedded between the second CrossStage module and the third CrossStage module of CSPDarkNet53, the backbone network of YOLOV4 model (that is, for each image frame, Need to run to the second CrossStage module of CSPDarkNet53 to aggregate the features, as shown in Fig. \ref{Aggregation mode 3}).

Because the bird's features are not obvious and its size is small in our detection task, the feature is easy to disappear in the process of feature extraction of a single frame image, and the deeper the layer is, the more obvious the feature disappears. Experimental results are shown in TABLE \ref{tab:Effect_Fea_Agg_Diff_loc}. Without Fine-Detection, the $\text{AP}_{50}$ of the aggregation mode I is 2.71\% higher than aggregation mode II and 28.38\% higher than aggregation mode III. The experimental results prove the above point of view, that is, for the flying bird object with unobvious features, the feature aggregation should be performed before the feature extraction as much as possible. At the same time, we can see that without Fine-Detection, the $\text{AP}_{50}$ of aggregation mode III is even 1.83\% lower than that of the still image based detection method YOLOV4. This indicates that feature aggregation, when the features of the flying bird object gradually disappear, will affect the feature expression of the flying bird object instead. Therefore, our algorithm adopts the aggregation mode I to aggregate the features of flying bird objects. In addition, for the same aggregation mode, the detection accuracy will be greatly improved after adding the Fine-Detection stage.

\begin{table}[!ht]
\caption{Effect of feature aggregation at different locations on the accuracy of the algorithm (FD represents Fine-Detection).\label{tab:Effect_Fea_Agg_Diff_loc}}
\centering
\begin{tabular}{c|c c c c}
\hline
Aggregation mode & $\text{AP}_{50}$ & $\text{AP}_{75}$  & AP & Run time(s) \\
\hline \hline
\textcolor[rgb]{0,1,0}{Mode I w/o FD} & \textcolor[rgb]{0,1,0}{0.6664} & \textcolor[rgb]{0,1,0}{0.0493} & \textcolor[rgb]{0,1,0}{0.2085} & \textcolor[rgb]{0,1,0}{0.0075} \\
Mode I & 0.7089 & 0.2011 & 0.3137 & 0.060 \\
\textcolor[rgb]{0,1,0}{Mode II w/o FD} & \textcolor[rgb]{0,1,0}{0.6393} & \textcolor[rgb]{0,1,0}{0.0488} & \textcolor[rgb]{0,1,0}{0.2011} & \textcolor[rgb]{0,1,0}{0.0094} \\
Mode II & 0.6830 & 0.1221 & 0.2734 & 0.085 \\
\textcolor[rgb]{0,1,0}{Mode III w/o FD} & \textcolor[rgb]{0,1,0}{0.3826} & \textcolor[rgb]{0,1,0}{0.0060} & \textcolor[rgb]{0,1,0}{0.0094} & \textcolor[rgb]{0,1,0}{0.014} \\
Mode III & 0.5778 & 0.1617 & 0.2489 & 0.090 \\
\hline
Method & $\text{AP}_{50}$ & $\text{AP}_{75}$  & AP & Run time(s) \\
\hline \hline
YOLOV4 & 0.4009 & 0.0719 & 0.1527 & 0.006 \\
\hline
\end{tabular}
\end{table}

\subsubsection{Effect of Different Number of Consecutive Input Frames on the Performance of the Algorithm.}

We design test experiments with different numbers of consecutive frame inputs to evaluate the impact on the detection accuracy and efficiency of the proposed method. Specifically, there are three consecutive frames of input, five consecutive frames of input, seven consecutive frames of input, etc. Theoretically, with the increase in the number of consecutive frames, the motion information of the flying bird object will be gradually enriched. However, the motion range of the flying bird object will also increase, and the difficulty of feature aggregation will also increase. Therefore, as the number of consecutive frames increases, the detection accuracy of the algorithm should be a process of first rising and then falling. Meanwhile, as the number of consecutive frames increases, the algorithm's running time will increase accordingly. The algorithm's detection performance test results are shown in TABLE \ref{tab:Effect_Diff_NUM} (the motion amount parameter ${\sigma}_{mov}$ is set to 4.0). The experimental results show that the algorithm has the fastest running speed but the lowest detection accuracy when input is input for three consecutive frames. The detection accuracy and running time are better when the input is for five consecutive frames. Therefore, five consecutive frames of input are used in our algorithm.

\begin{table}[!ht]
\caption{Effect of continuous image input with different number of frames on detection performance.\label{tab:Effect_Diff_NUM}}
\centering
\begin{tabular}{c|c c c c}
\hline
Frame num  & $\text{AP}_{50}$ & $\text{AP}_{75}$  & AP & Run Time(s) \\
\hline \hline
\textcolor[rgb]{0,1,0}{3 w/o FD}  & \textcolor[rgb]{0,1,0}{0.6409} & \textcolor[rgb]{0,1,0}{0.0277} & \textcolor[rgb]{0,1,0}{0.1971} & \textcolor[rgb]{0,1,0}{0.007} \\
3  & 0.6723 & 0.0821 & 0.2588 & 0.04 \\
\textcolor[rgb]{0,1,0}{5 w/o FD}  & \textcolor[rgb]{0,1,0}{0.6664} & \textcolor[rgb]{0,1,0}{0.0493} & \textcolor[rgb]{0,1,0}{0.2085} & \textcolor[rgb]{0,1,0}{0.0075} \\
5  & 0.7089 & 0.2011 & 0.3137 & 0.060 \\
\textcolor[rgb]{0,1,0}{7 w/o FD}  & \textcolor[rgb]{0,1,0}{0.6451} & \textcolor[rgb]{0,1,0}{0.0564} & \textcolor[rgb]{0,1,0}{0.2170} & \textcolor[rgb]{0,1,0}{0.0079} \\
7  & 0.7069 & 0.1359 & 0.2918 & 0.08 \\
\hline
\end{tabular}
\end{table}

\subsubsection{Influence of Different Amount of Motion Parameter ${\sigma}_{mov}$ on the Accuracy of the Algorithm}

\begin{figure}[!ht]
\centering
\includegraphics[width=3.2in]{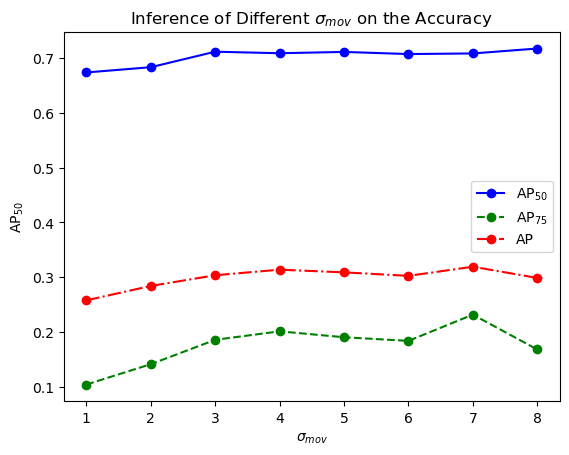}
\caption{Influence of Different Amount of Motion Parameter ${\sigma}_{mov}$ on the Accuracy of the Algorithm.}
\label{parameter_sigma_study_fig}
\end{figure}

We obtain the ASt-Cubes of different sizes of the flying bird object by setting different motion amount parameter ${\sigma}_{mov}$. If the MR is small, the context background information is less; if the MR is large, the SNR is large. Therefore, different sizes of MRs of the same flying bird object have different effects on the algorithm's performance. Fig. \ref{parameter_sigma_study_fig} shows the influence of different motion amount parameter ${\sigma}_{mov}$ on the algorithm's accuracy.

We can see from Fig. \ref{parameter_sigma_study_fig} that with the increase of parameter ${\sigma}_{mov}$, the detection precision of the proposed method has a first rise after the fall of a trend. As previously analyzed, when the MR is too small, it lacks contextual information, and when the MR is too large, it is easy to introduce more noise (in an extreme case, when the MR is already consistent with the original input image, the previous processing will be meaningless, because the input of the fine-detection stage is directly the original image. At the same time, because the fine-detection model is relatively simple and the input size is small (160 $\times$ 160), the detection effect is bound to be poor), so whether the MR is too small or too large will affect the accuracy of the algorithm. Through experiments, we find that when the motion amount parameter ${\sigma}_{mov}$ is 4.0, the detection performance of the algorithm is the best.

Through parameter analysis experiments, we conclude that when the number of consecutive input frames is 5, the algorithm can get a good balance between accuracy and speed. When the motion parameter is 4.0, the algorithm's accuracy reaches the highest. So we suggest the following parameter setting scheme. The number of consecutive input frames is set to 5, and the motion amount parameter ${\sigma}_{mov}$ is set to 4.0.

\subsection{Comparative Analysis Experiments}\label{comparative_analysis}

In order to verify the advancement of the proposed flying bird object detection algorithm. We design a series of comparative experiments to compare the accuracy of different methods in detecting flying bird objects. Specifically, we compare the static image object detection methods represented by YOLOV4 \cite{2020_Bochkovskiy_YOLOv4} and YOLOV5 \cite{yolov5_2021} and the video object detection methods represented by FGFA \cite{2017_Zhu_FGFA}, SELSA \cite{2019_Wu_SELSA}, and Temporal RoI Align \cite{2021_Tao_Temporal_RoI_Align}. These methods use the relevant open-source code, among which FGFA \cite{2017_Zhu_FGFA}, SELSA \cite{2019_Wu_SELSA}, and Temporal RoI Align \cite{2021_Tao_Temporal_RoI_Align} use the MMTracking \cite{mmtrack2020} open-source framework. Note that all methods are trained from scratch without pre-trained models.

By the way, the parameters of the proposed method are designed as follows. The input size is set to 640 $\times$ 384, the number of consecutive input frames is set to 5 frames,and the motion amount parameter ${\sigma}_{mov}$ is set to 4.0.

The results of quantitative comparison experiments are shown in TABLE \ref{tab:compare}. The experimental results show that for flying birds in surveillance videos, except for the method proposed in this paper, the YOLOV5 \cite{yolov5_2021} series achieves good results. Compared with YOLOV5l \cite{yolov5_2021}, the $\text{AP}_{50}$ of the proposed method is improved by 21.25\%, which proves once again that for flying bird objects with unobvious features and small size in a single frame image, their features are easy to disappear in the process of feature extraction, so it is necessary to perform feature aggregation for flying bird objects. Secondly, the detection accuracy of the video-based object detection method is not better than that of the static image-based detection method, and it is even lower than that of the advanced static image-based detection method (YOLOV5 \cite{yolov5_2021} series). Therefore, in the process of feature disappearance, feature aggregation will affect the feature expression of flying bird objects. Therefore, for flying bird objects with unobvious features and small sizes in a single frame image, their features should be aggregated before input into the model for feature extraction. In addition, the $\text{AP}_{75}$ of the proposed method is generally lower than that of the other methods, which indicates that the proposed method is less accurate than the other methods in detecting boxes.

\begin{table*}[!ht]
\caption{Comparison with other object detection methods.\label{tab:compare}}
\centering
\begin{tabular}{c|c|c c c}
\hline
Method  & backbnoe  & $\text{AP}_{50}$ & $\text{AP}_{75}$  & AP \\
\hline \hline
YOLOV4 \cite{2020_Bochkovskiy_YOLOv4} & CSPDarkNet53 & 0.4009 & 0.0719 & 0.1527 \\
YOLOV5s \cite{yolov5_2021} & CSPResNet50 & 0.4663 & - & 0.2492 \\
YOLOV5m \cite{yolov5_2021} & CSPResNet50 & 0.4950 & - & 0.2697 \\
YOLOV5l \cite{yolov5_2021} & CSPResNet50 & 0.4964 & - & 0.2867 \\
FGFA \cite{2017_Zhu_FGFA} & ResNet101 & 0.2729 & 0.1878 & 0.1586 \\
SELSA \cite{2019_Wu_SELSA} & ResNet101 & 0.4530 & 0.2688 & 0.2595 \\
Temporal RoI Align \cite{2021_Tao_Temporal_RoI_Align} & ResNet101 & 0.4424 & 0.2418 & 0.2331 \\
FBOD-BMI w/o FD & CSPDarkNet53 & 0.6664 & 0.0493 & 0.2085 \\
FBOD-BMI(ours) & CSPDarkNet53 & 0.7089 & 0.2011 & 0.3137 \\
\hline
\end{tabular}
\end{table*}

\begin{figure*}[!htp]
    \centering
    %第一行图片展示
    \subfloat{
        %左标题1
        \rotatebox{90}{\scriptsize{~~~~~~~Ground Truth}}
        \begin{minipage}[t]{0.225\linewidth}
	\centering
	\includegraphics[width=1\linewidth]{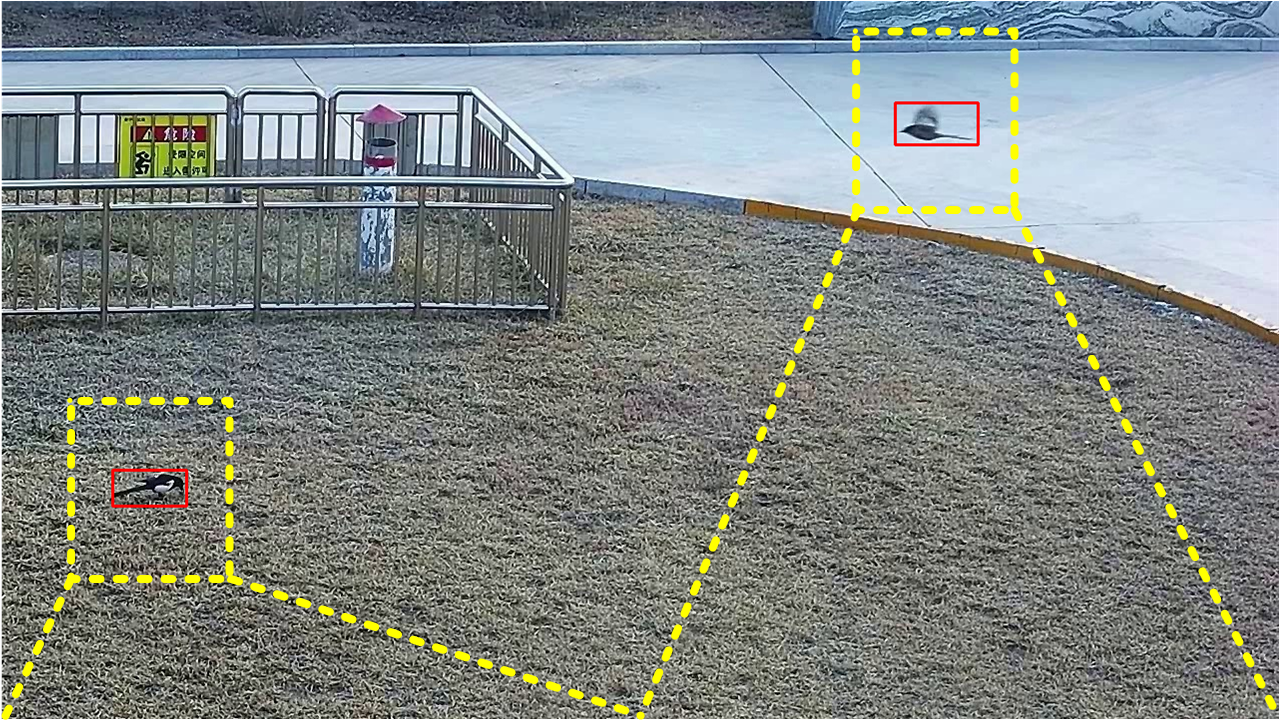}
	\end{minipage}
	}
    \subfloat{
	\begin{minipage}[t]{0.225\linewidth}
	\centering
	\includegraphics[width=1\linewidth]{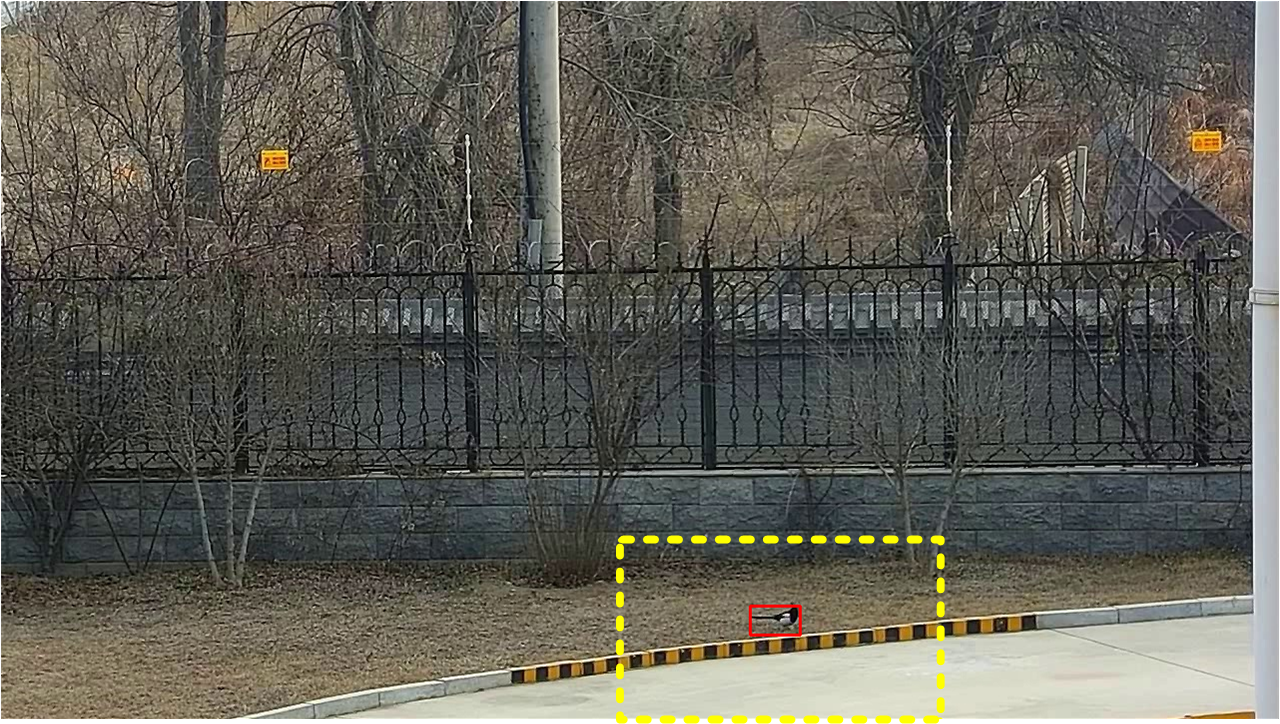}
	\end{minipage}
	}
    \subfloat{
        \begin{minipage}[t]{0.225\linewidth}
        \centering
        \includegraphics[width=1\linewidth]{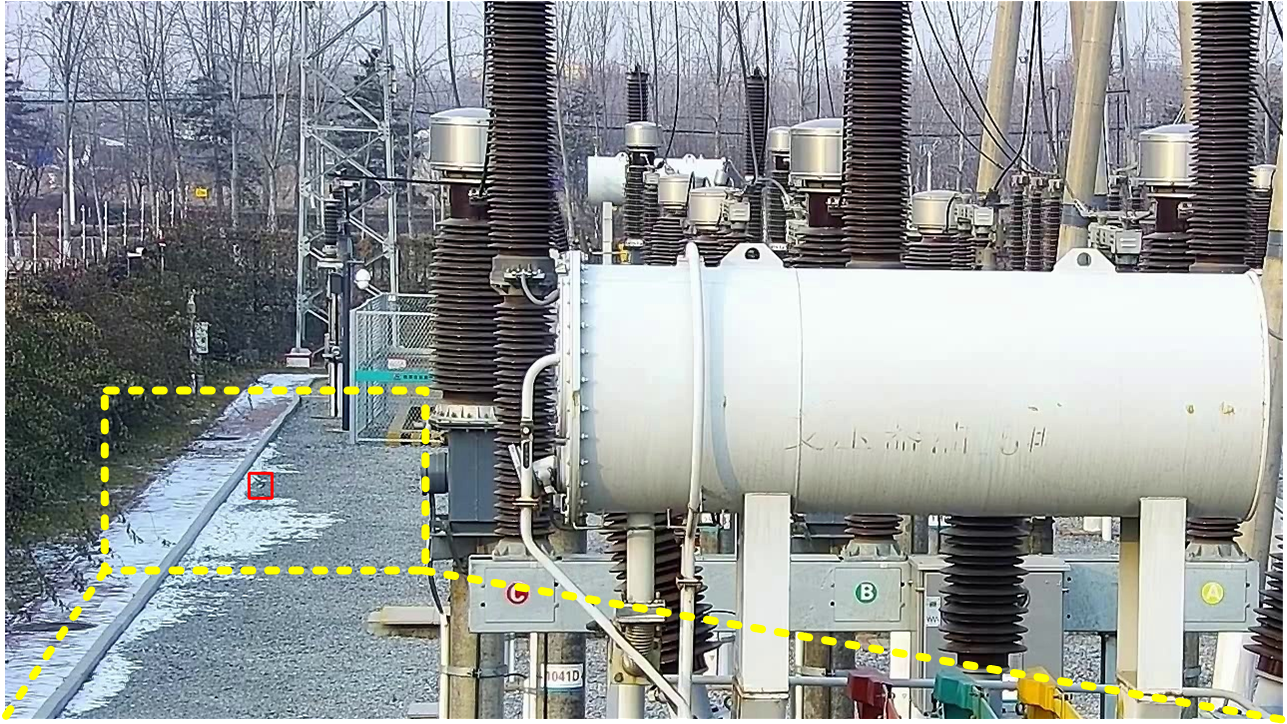}
        \end{minipage}
        }
    \subfloat{
        \begin{minipage}[t]{0.225\linewidth}
        \centering
        \includegraphics[width=1\linewidth]{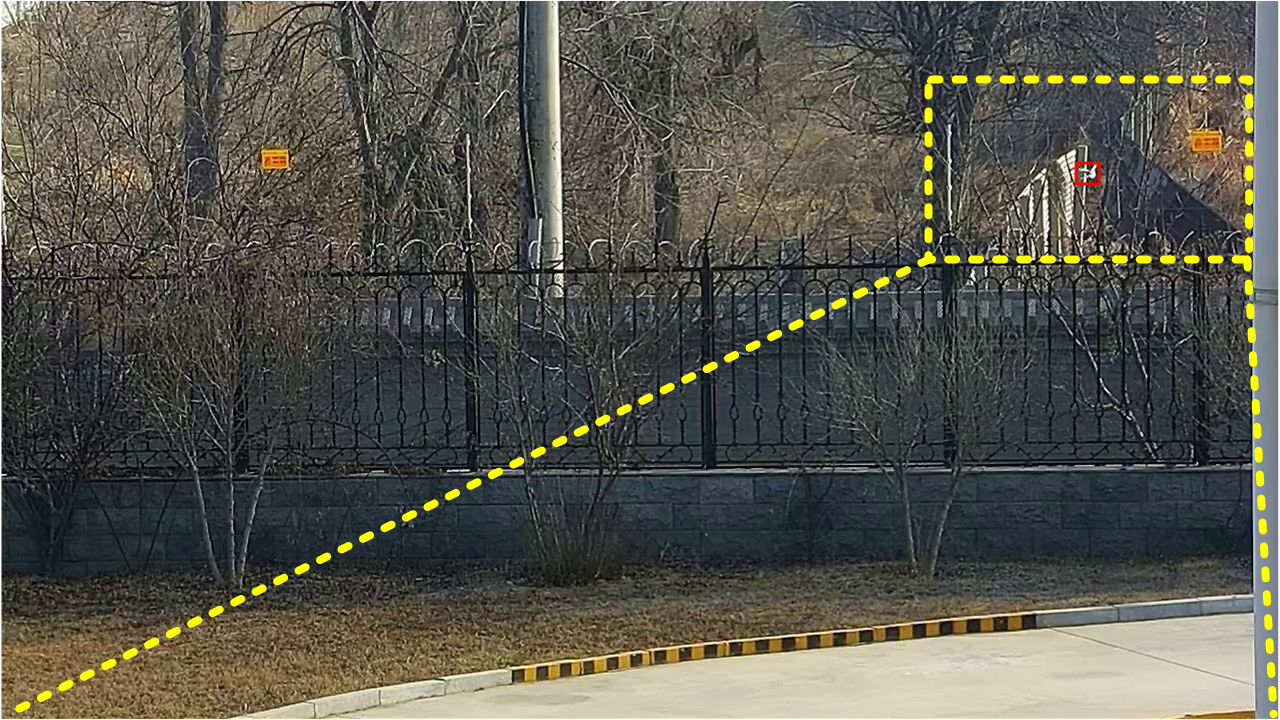}
        \end{minipage}
        }
    % 两行图片的间隙有点大，通过vspace进行微调
    \vspace{-2mm}
    % 由于上面已经用了subfigure，下面我们希望从 a 重新编号，而不是从 d 开始，清零。
    \setcounter{subfigure}{0}
    % 第二行图片展示
    \subfloat{
        % 左标题2
	\rotatebox{90}{\scriptsize{~~~~~~~~~YOLOV5l}}
	\begin{minipage}[t]{0.225\linewidth}
	\centering
	\includegraphics[width=1\linewidth]{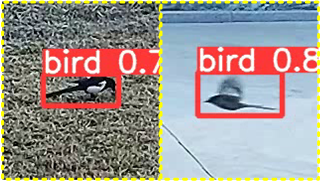}
	\end{minipage}
	}
    \subfloat{
	\begin{minipage}[t]{0.225\linewidth}
	\centering
	\includegraphics[width=1\linewidth]{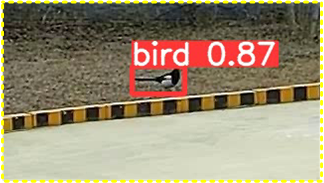}
	\end{minipage}
	}
    \subfloat{
	\begin{minipage}[t]{0.225\linewidth}
	\centering
	\includegraphics[width=1\linewidth]{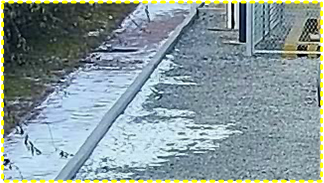}
	\end{minipage}
	}
    \subfloat{
	\begin{minipage}[t]{0.225\linewidth}
	\centering
	\includegraphics[width=1\linewidth]{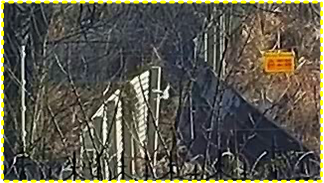}
	\end{minipage}
	}
    % 两行图片的间隙有点大，通过vspace进行微调
    \vspace{-2mm}
    % 由于上面已经用了subfigure，下面我们希望从 a 重新编号，而不是从 d 开始，清零。
    \setcounter{subfigure}{0}
    % 第三行图片展示
    \subfloat{
        % 左标题2
        \rotatebox{90}{\scriptsize{~~~~~~~~~~SELSA}}
        \begin{minipage}[t]{0.225\linewidth}
	\centering
	\includegraphics[width=1\linewidth]{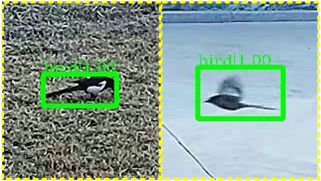}
	\end{minipage}
	}
    \subfloat{
	\begin{minipage}[t]{0.225\linewidth}
	\centering
	\includegraphics[width=1\linewidth]{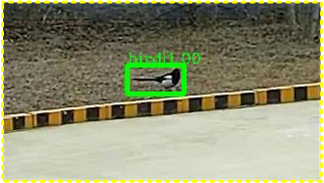}
	\end{minipage}
	}
    \subfloat{
	\begin{minipage}[t]{0.225\linewidth}
	\centering
	\includegraphics[width=1\linewidth]{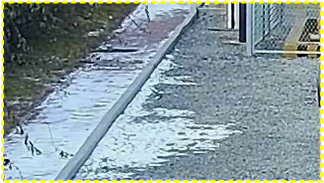}
	\end{minipage}
	}
    \subfloat{
	\begin{minipage}[t]{0.225\linewidth}
	\centering
	\includegraphics[width=1\linewidth]{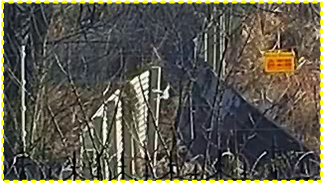}
	\end{minipage}
	}
    % 两行图片的间隙有点大，通过vspace进行微调
    \vspace{-2mm}
    % 由于上面已经用了subfigure，下面我们希望从 a 重新编号，而不是从 d 开始，清零。
    \setcounter{subfigure}{0}
    % 第四行图片展示
    \subfloat{
        % 左标题2
        \rotatebox{90}{\scriptsize{~FBOD-BMI w/o FD}}
        \begin{minipage}[t]{0.225\linewidth}
	\centering
	\includegraphics[width=1\linewidth]{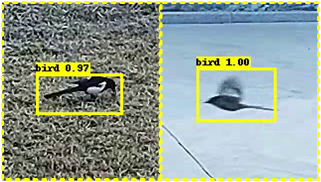}
	\end{minipage}
	}
    \subfloat{
	\begin{minipage}[t]{0.225\linewidth}
	\centering
	\includegraphics[width=1\linewidth]{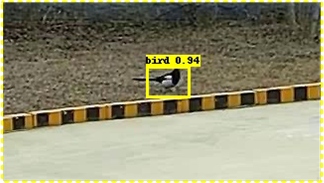}
	\end{minipage}
	}
    \subfloat{
	\begin{minipage}[t]{0.225\linewidth}
	\centering
	\includegraphics[width=1\linewidth]{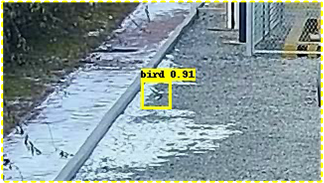}
	\end{minipage}
	}
    \subfloat{
	\begin{minipage}[t]{0.225\linewidth}
	\centering
	\includegraphics[width=1\linewidth]{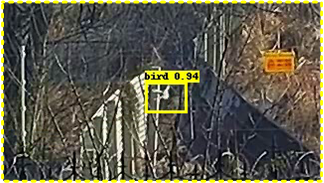}
	\end{minipage}
	}
    % 两行图片的间隙有点大，通过vspace进行微调
    \vspace{-2mm}
    % 由于上面已经用了subfigure，下面我们希望从 a 重新编号，而不是从 d 开始，清零。
    \setcounter{subfigure}{0}
    % 第五行图片展示
    \subfloat[Scenario 1]{
        % 左标题2
	\rotatebox{90}{\scriptsize{~~~~FBOD-BMI}}
	\begin{minipage}[t]{0.225\linewidth}
	\centering
	\includegraphics[width=1\linewidth]{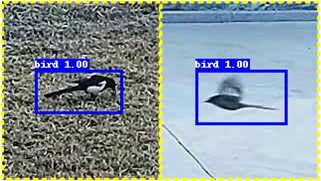}
	\label{scenario_1}
	\end{minipage}
	}
    \subfloat[Scenario 2]{
	\begin{minipage}[t]{0.225\linewidth}
	\centering
	\includegraphics[width=1\linewidth]{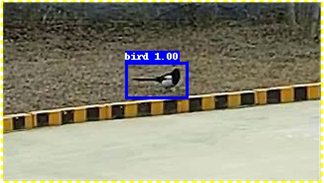}
	\label{scenario_2}
	\end{minipage}
	}
    \subfloat[Scenario 3]{
	\begin{minipage}[t]{0.225\linewidth}
	\centering
	\includegraphics[width=1\linewidth]{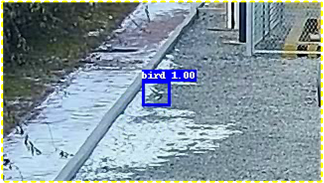}
	\label{scenario_3}
	\end{minipage}
	}
    \subfloat[Scenario 4]{
	\begin{minipage}[t]{0.225\linewidth}
	\centering
	\includegraphics[width=1\linewidth]{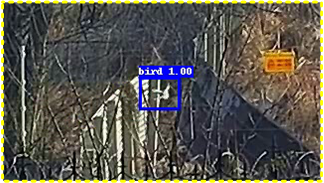}
	\label{scenario_4}
	\end{minipage}
	}
 
	% 添加题注，即对这个图片的说明
	\caption{Comparison of detection effect of different methods. The birds in Scenario 1 and 2 are clear. The birds in Scenario 3 and 4 are blurred and have indistinct features.}
	\label{Detection_results_on_BirdDataset}
\end{figure*}

In the qualitative comparison experiment, YOLOV5l \cite{yolov5_2021} and SELSA \cite{2019_Wu_SELSA} with higher detection accuracy are selected to compare with the proposed method. By comparing the experimental results shown in Fig. \ref{Detection_results_on_BirdDataset}, it can be seen that when the appearance characteristics of the flying bird object are relatively obvious, all the methods can detect the object (see Fig. \ref{scenario_1} and Fig. \ref{scenario_2}). However, YOLOV5l \cite{yolov5_2021} and SELSA \cite{2019_Wu_SELSA} will miss detection when the appearance features of a single frame image of a flying bird object are not obvious. However, the proposed method can achieve good results regardless of whether the appearance features of the single frame image of the flying bird object are obvious or not (see Fig. \ref{scenario_3} and Fig. \ref{scenario_4}). At the same time, we find that the detection box of the proposed method is not as compact as YOLOV5l \cite{yolov5_2021} and SELSA \cite{2019_Wu_SELSA}, which verifies the experimental results in the table, that is, with the increase of IOU threshold, the detection accuracy of the proposed method decreases faster than that of YOLOV5 \cite{yolov5_2021} and SELSA \cite{2019_Wu_SELSA}. However, this does not affect the effectiveness of the proposed method because for our task, detection is more important than detection accuracy, and since the object is small, such a difference can be ignored.

Through the qualitative and quantitative analysis of the experimental results, the flying bird object detection method proposed in this paper is advanced and effective.

% ===================================================================================================================================
% ===================================================================================================================================

% Note that IEEE does not put floats in the very first column - or typically
% anywhere on the first page for that matter. Also, in-text middle ("here")
% positioning is not used. Most IEEE journals use top floats exclusively.
% Note that, LaTeX2e, unlike IEEE journals, places footnotes above bottom
% floats. This can be corrected via the \fnbelowfloat command of the
% stfloats package.

\section{Conclusion}\label{Conclusion}

This paper proposes a flying bird Object Detection algorithm Based on Motion Information (FBOD-BMI) to solve the problem that the features of the object are not obvious in a single frame and the object's size is small (low SNR in surveillance video). Firstly, the ConvLSTM-PANet model was used to coarse detect the whole frame of continuous video frames to capture suspicious flying bird objects. Then, the object tracking method tracks the suspicious flying bird object, and the MR Of the suspicious flying bird object on $n$ consecutive frames of images is determined. At the same time, according to the movement speed of the suspicious flying bird object, the size of its MR is adaptively adjusted (specifically, if the object is moving slowly, its MR is expanded according to its speed to ensure the context environment information), and its adaptive St-Cubes (ASt-Cubes) are generated to ensure that the SNR of the flying bird object is improved while the necessary environmental information is retained adaptively. Then, the ASt-Cubes of suspicious flying bird objects are accurately classified and located by the LW-USN model. Finally, qualitative and quantitative experiments verify the effectiveness of the proposed flying bird object detection algorithm based on motion information. At the same time, we proved through ablation experiments that for the problem that the features of single frame images of flying bird objects in videos are not obvious, it is necessary to perform feature fusion before feature extraction as much as possible.

% or


% ============================================
%\appendices
%Appendix one text goes here %\cite{Roberg2010}.

%\section{}

%\section*{Acknowledgment}

%The authors would like to thank D. Root for the loan of the SWAP. The SWAP that can ONLY be usefull in Boulder...

\ifCLASSOPTIONcaptionsoff
  \newpage
\fi



%\begin{thebibliography}{1}


\bibliographystyle{IEEEtran}
\bibliography{IEEEabrv,Bibliography}

\vfill


\end{document}